\documentclass[twoside,11pt]{article}

\usepackage{jmlr2e}

\usepackage{times}
\usepackage{epsfig}
\usepackage{graphicx}
\usepackage{amssymb}
\usepackage{color}
\usepackage{dsfont}
\usepackage[algo2e]{algorithm2e}
\usepackage{algorithmic}
\usepackage{algorithm}
\usepackage{subfigure}
\usepackage{booktabs}
\usepackage{multirow}
\usepackage{amsmath}
\usepackage{hyperref}

\def \bx{\mathbf{x}}

\def \bX{\mathbf{X}}

\def \bz{\mathbf{z}}

\def \bOmega{\mathbf{\Omega}}

\def \bW{\mathbf{W}}

\def \bb{\mathbf{b}}

\def \cL{\mathcal{L}}

\def \bh{\mathbf{h}}
\def \bx{\mathbf{x}}
\def \bX{\mathbf{X}}

\def \bOmega{\mathbf{\Omega}}
\def \bW{\mathbf{W}}
\def \bb{\mathbf{b}}

\def \cI{\mathcal{I}}
\def \cL{\mathcal{L}}
\def \bz{\mathbf{z}}
\def \bp{\mathbf{p}}
\newtheorem{ap_lemma}{Lemma}
\newtheorem{ap_definition}{Definition}

\usepackage{lastpage}
\jmlrheading{22}{2021}{1-\pageref{LastPage}}{6/19; Revised
9/21}{9/21}{19-497}{Xi Peng, Yunfan Li, Ivor Tsang, Hongyuan Zhu,  Jiancheng Lv, and Joey Tianyi Zhou}

\ShortHeadings{XAI Beyond Classification: Interpretable Neural Clustering}{Peng, Li, Tsang, Zhu, Lv, and Zhou.}

\firstpageno{1}

\begin{document}

\title{XAI Beyond Classification: Interpretable Neural Clustering}


\author{\name Xi Peng$^1$ 
\email pengx.gm@gmail.com 
\AND
\name Yunfan Li$^1$ 
\email yunfanli.gm@gmail.com 
\AND
\name Ivor W. Tsang$^{2,3}$
\email ivor.tsang@gmail.com
\AND
\name Hongyuan Zhu$^4$
\email zhuh@i2r.a-star.edu.sg
\AND
\name Jiancheng Lv$^1$\thanks{Corresponding author.}
\email lvjiancheng@scu.edu.cn
\AND
\name Joey Tianyi Zhou$^{3,5}$
\email joey.tianyi.zhou@gmail.com\\
\addr $^1$College of Computer Science, Sichuan University, Chengdu, China.\\
\addr $^2$Centre for Frontier Artificial Intelligence Research, A*STAR, Singapore.\\
\addr $^3$Australian Artificial Intelligence Institute, University of Technology, Sydney Australia.\\
\addr $^4$Institute for Infocomm Research, A*STAR, Singapore.\\
\addr $^5$Institute of High Performance Computing, A*STAR, Singapore.\\
}

\editor{David Blei}

\maketitle

\begin{abstract}
In this paper, we study two challenging problems in explainable AI (XAI) and data clustering. The first is how to directly design a neural network with inherent interpretability, rather than giving post-hoc explanations of a black-box model. The second is implementing discrete $k$-means with a differentiable neural network that embraces the advantages of parallel computing, online clustering, and clustering-favorable representation learning. To address these two challenges, we design a novel neural network, which is a differentiable reformulation of the vanilla $k$-means, called inTerpretable nEuraL cLustering (TELL). Our contributions are threefold. First, to the best of our knowledge, most existing XAI works focus on supervised learning paradigms. This work is one of the few XAI studies on unsupervised learning, in particular, data clustering. Second, TELL is an interpretable, or the so-called intrinsically explainable and transparent model. In contrast, most existing XAI studies resort to various means for understanding a black-box model with post-hoc explanations. Third, from the view of data clustering, TELL possesses many properties highly desired by $k$-means, including but not limited to online clustering, plug-and-play module, parallel computing, and provable convergence. Extensive experiments show that our method achieves superior performance comparing with 14 clustering approaches on three challenging data sets. The source code could be accessed at \url{www.pengxi.me}.
\end{abstract}


\section{Introduction}
\label{sec1}

As a fundamental topic in machine learning, clustering aims to group similar samples into the same cluster and separate dissimilar ones into different clusters. During the past decade, a variety of clustering methods~\citep{Jain1999:Survey} have been proposed and achieved encouraging success in various applications. In recent, the main focus of the community shifts to how to handle high-dimensional data that is usually linear inseparable.

To effectively cluster high-dimensional data, many kinds of methods have been proposed, \textit{e.g.}, spectral clustering~\citep{Ng2002}, kernel clustering~\citep{JMLR:v20:17-517}, convex clustering~\citep{Hocking:2011tj,Zhanglijun:2013}, subspace clustering~\citep{Elhamifar2013,Liu2013,Lu2012,Yang2018:IJCV,Li2015:Clustering}, and the recent popular deep  clustering~\citep{Yang:2016kn,Peng2016:Deep_abbv,Ji2017:Deep}. The aforementioned methods share a common clustering paradigm of first learning a shallow or deep representation and then applying a traditional clustering method  ($k$-means in most cases) to make cluster assignments.

Though promising results have been achieved on many applications, these methods still suffer from the following limitations. Namely, though grounded in theory, traditional approaches such as subspace clustering might be incapable of handling more complex data due to its limited representability. On the contrary, although deep clustering methods could capture the hidden nonlinear structure of data, as a ``black box'' model, their lack of explainability makes its working mechanism hard to understand. Consequently, unguided and laborious hyper-parameter tuning is usually required to achieve satisfying results.

\begin{figure}[!t]
\centering
\includegraphics[width=0.8\textwidth]{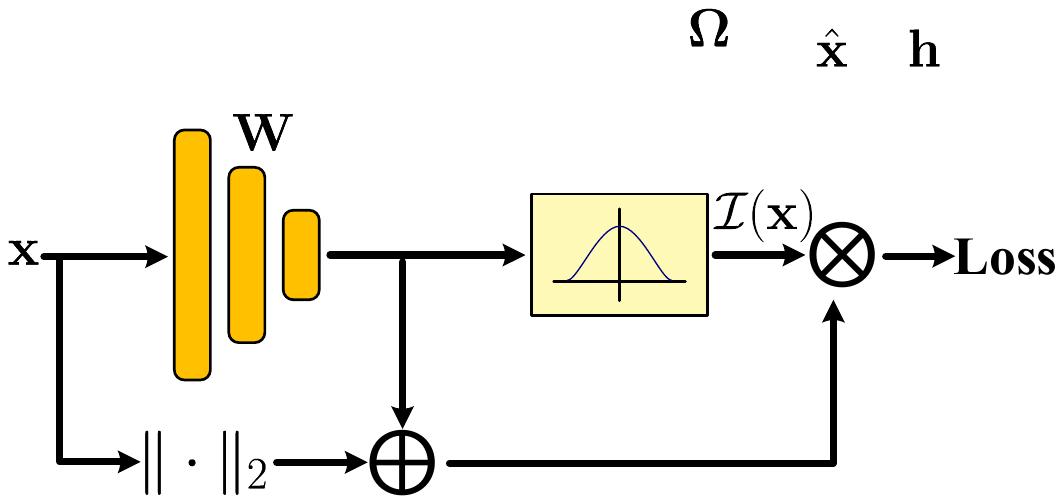}
\caption{\label{fig1} An illustration of the proposed TELL. This  structure is exactly derived from the vanilla $k$-means objective and exhibits explicit interpretability. In brief, the hyperplane $\bW$ is spanned by the clustering centers $\bOmega$ and the activation function normalizes the cluster assignment $\mathcal{I}(\cdot)$, which enjoys the model decomposability and algorithmic transparency as discussed in Section~\ref{sec:3.3}.}
\end{figure}

In this paper, we propose a novel neural network (illustrated in Fig.~\ref{fig1}) from the perspective of differentiable programming (DP) and learning-based optimization~\citep{Gregor:2010, Sprechmann2015:DeepRank, Zheng2015:CRFRNN, Liu2016:PAMI, Chen:2018vc, Liu:2019ta, Long:2018uy}. The proposed inTerpretable nEuraL cLustering (TELL) is a differentiable alternative of the vanilla $k$-means, which reformulates the $k$-means objective as a neural layer. As a differentiable reformulation, TELL equips the vanilla $k$-means with advantages of neural networks, including end-to-end optimization, pluggability, provable convergence, and interpretable working mechanism. It could achieve clustering for large-scale and online data, which is impractical for the vanilla $k$-means.

The contribution and novelty of this work are summarized as follows:
\begin{itemize}
	\item From the view of XAI, our contribution is twofold. On the one hand, we directly build an interpretable neural network rather than design some post-hoc analyses to explain a neural network like most existing XAI works did. As pointed out by~\citep{Rudin:2019ha}, a large number of works have been conducted on the explainability of black-box models, but few efforts have been made on directly building an interpretable model. This work could be a valuable attempt towards this direction. On the other hand, most existing interpretable neural networks like the well-known  perceptron~\citep{Rosenblatt1961, Freund1999} are designed for supervised tasks. To the best of our knowledge, this could be the first attempt on interpretable neural networks for unsupervised tasks, or more specifically, clustering in this work.
	\item From the view of clustering, TELL implements the vanilla $k$-means with a neural network by reformulating its discrete objective as a neural layer, which enjoys the following advantages. First, the proposed TELL could be easily optimized by SGD in parallel, and we theoretically prove that the loss could be monotonously reduced. Second, the vanilla $k$-means requires the entire data set to update cluster centers in each iteration, which is computationally inefficient for large-scale data and even incapable of handling online data, \textit{i.e.}, the data presented in streams. In contrast, our TELL optimizes the cluster centers through batch-wise SGD and directly predicts the cluster assignment for each point, which is promising in clustering large-scale and online data. Third, different from the vanilla $k$-means, TELL can be plugged into any neural network to help it learning a clustering-favorable representation in an end-to-end fashion.
	\item From the view of differentiable programming, as far as we know, this could be one of the first attempts to benefit clustering with DP. On the one hand, this work aims at differentiable data clustering, whereas most existing DP works only focus on solving an optimization problem using a neural network~\citep{Gregor:2010, Liu2016:PAMI, Zuo:2015hh, Chen2015:On}. On the other hand, our TELL recasts the vanilla $k$-means as a one-layer feedforward neural network (FNN), whereas most existing DP methods~\citep{Wang2015:Learning, Sprechmann2015:DeepRank, Liu2018:TNNLS} are build on recurrent neural networks (RNN). Therefore, this work might provide some novel insights to the community.
\end{itemize}

\noindent \textit{Mathematical Notations:} Throughout the paper, \textbf{lower-case bold letters} represent column vectors and \textbf{upper-case bold letters} denote matrices. $\mathbf{A}^{\top}$ denotes the transpose of the matrix $\mathbf{A}$ and $\mathbf{I}$ denotes the identity matrix.

\section{Interpretable Neural Clustering}
\label{sec3}

In this section, we first show how to recast the vanilla $k$-means objective to a differentiable one on which a neural layer is built. Then, we discuss the interpretability of our model from the perspective of XAI, followed by the convergence proofs.

\subsection{Deficiency of the Vanilla $k$-means}
\label{sec:3.1}

For a given data set $\bX=\{\bX_{1},\bX_{2},\cdots,\bX_{n}\}$, $k$-means aims to group each point $\bX_{i}$ into one of $k\leq n$ sets $\mathcal{S}=\{\mathcal{S}_{1},\mathcal{S}_{2},\cdots,\mathcal{S}_{k}\}$ by minimizing the distance of the within-cluster data points, \textit{i.e.},
\begin{equation}
	\label{eq3.1}
	\underset{\mathcal{S}}{\text{argmin}}\sum_{j}\sum_{\bX_{i}\in\mathcal{S}_{j}}\|\bX_{i} - \bOmega_{j} \|_{2}^{2},
\end{equation}
where $\mathbf{\bOmega}_{j}$ denotes the $j$-th cluster center which is computed as the mean of points in $\mathcal{S}_{j}$, \textit{i.e.},
\begin{equation}
\label{eq3.2}
	\mathbf{\bOmega}_{j} = \frac{1}{|\mathcal{S}_{j}|}\sum_{\bX_{i}\in \mathcal{S}_{j}}\bX_{i},
\end{equation}
where $|\mathcal{S}_{j}|$ denotes the number of data points in the $j$-th cluster.

To solve Eq.~(\ref{eq3.1}), an EM-like optimization is adopted by updating $\mathcal{S}$ and $\bOmega$ iteratively, \textit{i.e.}, fixing one while optimizing the other. Such an iterative optimization has several drawbacks.

First, it is NP-hard to find the optimal solution for $k$-means in the Euclidean space, even for the bi-cluster problem. To ease the NP-hard problem, some variants of $k$-means are proposed, \textit{e.g.}, parametric methods like Fuzzy $c$-means~\citep{Dunn1973:FCM, Bezdek1981:FCM}. However, these methods are sensitive to the value of hyper-parameters that are daunting to tune.

Second, the vanilla $k$-means requires the entire data set to compute the cluster centers in each iteration~\citep{Yang2018:IJCV, NIPS2019_8741, NIPS2019_9442}. As a result, it is impractical in large-scale or online clustering scenario, where data is presented in streams. More precisely, although one could assign the new-coming data to its nearest cluster center, the centers cannot be further updated unless one replicates the algorithm on all data, including the old and new.

Third, the vanilla $k$-means is conducted on the fixed inputs and cannot assist the representation learning. As the success of deep learning largely depends on end-to-end learning, a plug-and-play neural clustering module is highly expected. In the proposed TELL, the cluster layer could not only perform clustering but also help the network to learn clustering-favorable representations in an end-to-end manner.

\subsection{The Proposed Method}

\label{sec:3.2}
In this section, we first elaborate on how to reformulate the vanilla $k$-means into a differentiable neural layer in Section~\ref{sec:3.2.1}. Then, in Section~\ref{sec:3.2.2}, we theoretically prove the necessity of decoupling the weight $\bW$ and bias $\bb$ of the neural layer even though they are inherently correlated. In Section~\ref{sec:3.2.3}, we further reveal that such a decoupling strategy could cause divergent and unstable training. As a solution, we propose normalizing both the weight of the cluster layer and its gradient to stabilize the training. Finally, we present an end-to-end framework that could simultaneously learn a clustering-favorable representation and achieve clustering in Section~\ref{sec:3.2.4}, which proves the plug-and-play characteristic and effectiveness of the proposed TELL.

\subsubsection{Neural Network Implementation of $k$-means}
\label{sec:3.2.1}
To overcome drawbacks of the vanilla $k$-means mentioned in Section~\ref{sec:3.1}, we recast its objective function into a neural layer by rewriting Eq.~(\ref{eq3.1}) into
\begin{equation}
\label{eq3.3}
	\min \sum_{i=1}^{n}\sum_{j=1}^{k} \cI_{j}(\bX_{i})\|\bX_{i}-\bOmega_{j}\|_{2}^{2},
\end{equation}
where $\cI_{j}(\bX_{i})$ indicates the cluster membership of $\bX_{i}$ \textit{w.r.t.} $\bOmega_{j}$ and only one entry of $\cI_{j}(\bX_{i})$ is nonzero. 

The right part of Eq.~(\ref{eq3.3}) could be expanded into
\begin{equation}
\label{eq.3.4b}
	\|\bX_{i}-\bOmega_{j} \|_{2}^{2}=\|\bX_{i}\|_{2}^{2}-2\bOmega_{j}^{\top}\bX_{i}+\|\bOmega_{j}\|_{2}^{2}.
\end{equation}

We then define
\begin{equation}
\label{eq.3.4c}
	\bW_j=2\bOmega_j, \hspace{3mm}\bb_{j}=-\|\bOmega_{j}\|_{2}^{2}, \hspace{3mm}\|\bX_{i}\|_{2}^{2}=\beta_{i}\geq 0,
\end{equation}
where $\bW_j$ is the $j$-th column of $\bW$, $\bb_{j}$ is a scalar which denotes the $j$-th entry of $\bb$, and $\beta_i$ is a nonnegative constant corresponding to the length of data point $\bX_{i}$. 

With the above formulations, we could equivalently recast the scatter between data point $\bX_{i}$ and cluster center $\bOmega_{j}$ as
\begin{equation}
\label{eq:3.4}
    \|\bX_{i}-\bOmega_{j}\|_{2}^{2} = \beta_{i}-\bW_{j}^{\top}\bX_{i} - \bb_{j}. 
\end{equation}

For a given temperature factor $\tau>0$, we relax the categorical variable $\cI_{j}(\bX_{i})$ into
\begin{equation}
	\label{eq3.5}
	\cI_{j}(\bX_{i}) = \frac{\exp(- \|\bX_{i}-\bOmega_{j}\|_{2}^{2}~/\tau)}{\sum_{k}\exp(-\|\bX_{i}-\bOmega_{k}\|_{2}^{2}~/\tau)}.
\end{equation}

In fact, the above definition of $\cI_{j}(\bX_{i})$ can be regarded as the attention of $\bX_{i}$ on the $j$-th cluster, which will be elaborated later in Section~\ref{sec:3.3}.

Combining Eq.~(\ref{eq:3.4}) and Eq.~(\ref{eq3.5}), $\cI_{j}(\bX_{i})$ could be computed with the proposed neural layer through
\begin{equation}
	\label{eq3.6}
	\cI_{j}(\bX_{i})=\frac{\exp((\bW^{\top}_{j}\bX_{i}+\bb_{j}-\beta_{i})/\tau)}{\sum_{k}\exp((\bW^{\top}_{k}\bX_{i}+\bb_{k}-\beta_{k})/\tau)}.
\end{equation}

Notably, the continuous categorical variable $\cI_{j}(\bX_{i})$ could be computed using any normalization function including but not limited to \textit{softmax} here. To avoid exhaustively  tuning on the temperature parameter, in our implementation, we adopt an alternative by simply keeping the maximal entry of $\cI_{j}(\bX_{i})$, which is the case when $\tau$ approaches 0 and is consistent with the vanilla $k$-means.

\subsubsection{Decoupling the Network Weight and Bias}
\label{sec:3.2.2}

To avoid confusions brought by complex mathematical notations, in the following analysis, we simply consider the case of one sample $\bx$ without loss of generality. In this case, the objective function of TELL could be formulated as
\begin{equation}
	\label{eq3.7}
	\cL=\sum_{j}\cL_{j} = \sum_{j}\cI_{j}(-\bW_{j}^{\top}\bx - \bb_{j}+\beta),
\end{equation}
where $\cI_{j}$ shorts for $\cI_{j}(\bx)$. 
  
Though $\bW$ and $\bb$ are inherently coupled (\textit{i.e.}, $\bb_j=-\frac{\|\bW_j\|_{2}^{2}}{4}$) according to the definition in Eq.~(\ref{eq.3.4c}), we theoretically prove that $\bW$ and $\bb$ should be decoupled during the training. In other words, $\bW$ and $\bb$ are optimized independently and the final cluster centers $\bOmega^{\ast}$ are recovered via $\bOmega^{\ast}=\frac{1}{2}\bW^{\ast}$.

To demonstrate the necessity of decoupling $\bW$ and $\bb$, similar to the above reformulation, we rewrite Eq.~(\ref{eq3.7}) into
\begin{align}
	\label{eq3.8}
	\cL&=-\sum_{j}\frac{\exp((\bb_{j}+\bW_{j}^{\top}\bx-\beta)/\tau)(\bb_{j}+\bW_{j}^{\top}\bx-\beta)}{\sum_{k}\exp((\bb_{k}+\bW_{k}^{\top}\bx-\beta)/\tau)}\notag\\
	&=-\sum_{j}\frac{\exp(\bz_{j}/\tau)}{\sum_{k}\exp(\bz_{k}/\tau)}\bz_{j}\notag\\
	&=-\sum_{j}f(\bz_{j}),
\end{align}
where $\bz_{j}=(-\frac{\|\bW_{j}\|_{2}^{2}}{4}+\bW_{j}^{\top}\bx-\beta)$.

Correspondingly, the objective function becomes
\begin{equation}
\begin{aligned}
\label{eq3.9}
	\max~& \sum_{j}f(\bz_{j}) \\
	\text{s.t.}~& \bz_{j}=-\frac{\|\bW_{j}\|_{2}^{2}}{4}+\bW_{j}^{\top}\bx-\beta.
\end{aligned}
\end{equation}

As can be seen, Eq.~(\ref{eq3.9}) is equivalent to Eq.~(3) when $\bW$ and $\bb$ are coupled, \textit{i.e.}, $\bb_j=-\frac{\|\bW_j\|_{2}^{2}}{4}$. Since Eq.~(\ref{eq3.9}) obtains the optimum at the boundary with $\bz_{1}=\bz_{2}=\cdots$ and $f(\bz)=\infty$ when $\bz=\infty$, there exists $\bz^{\ast}$ such that $\bz_{j}=\bz^{\ast}$ and $f(\bz^{\ast})$ reaches the optimum. We can always find a $\bW_{j}$ and $\bb_{j}$ such that $\bb_{j}+\bW_{j}^{\top}\bx-\beta=\bz^{\ast}$, while it is not guaranteed to have a $\bW_{j}$ and $\bb_{j}$ such that $-\frac{\|\bW_{j}\|_{2}^{2}}{4}+\bW_{j}^{\top}\bx-\beta=\bz^{\ast}$. Notably, though the above analyses are based on the case of a single sample, the conclusion still holds for multiple samples since they are independent from each other. In this sense, we have to decouple $\bW_{j}$ and $\bb_{j}$ during training to avoid the trivial solution.

\subsubsection{Normalize the Cluster Layer Weight and Gradient}
\label{sec:3.2.3}

\begin{figure}[!t]
\centering
\includegraphics[width=0.8\textwidth]{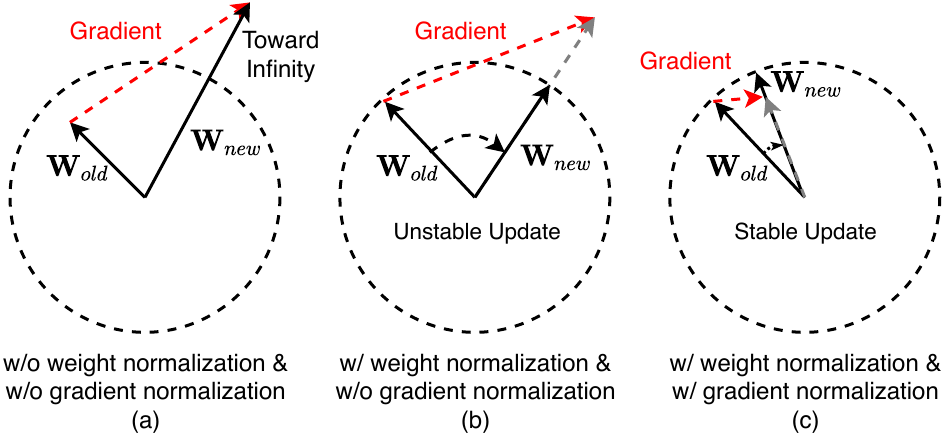}
\caption{\label{fig:normalization} Three different cases of optimizing the cluster layer $\bW$, \textit{i.e.}, updating the cluster centers $\bOmega$. (a) Directly minimizing $-\bW^\top\bX$ with SGD, which leads $\bW$ to infinity and non-convergence. (b) Conducting weight normalization after each SGD update. This will prevent $\bW$ from going to infinity. However, when the gradient is much larger than the weight, $\bW$ will be greatly changed after each update, which ends up in unstable training because the physical meaning of $\bW$ may differ in each iteration. (c) Conducting gradient normalization and weight normalization, which promises stable training and is adopted in our implementation.
}
\end{figure}

In Section \ref{sec:3.2.2}, we have shown that it is necessary to decouple $\bW$ and $\bb$ for preventing the network from descending into a trivial solution. However, we further notice that when $\bW$ and $\bb$ are decoupled, directly optimizing them would lead to divergent and unstable training. To address this issue, we propose normalizing both the cluster layer weight and its gradient to achieve a stable training, as illustrated in Fig.~\ref{fig:normalization}.

To be specific, when $\bW$ and $\bb$ are decoupled, minimizing the loss function $\sum_{j}\cI_{j}(-\bW_{j}^{\top}\bx - \bb_{j}+\beta)$ in Eq.~(\ref{eq3.7}) would lead both $\bW_{j}^{\top}$ and $\bb_{j}$ to infinity, as shown in Fig.~\ref{fig:normalization}(a). In this case, the optimization of the cluster layer never converges. To solve this problem, we propose simultaneously normalizing the weight and bias of the cluster layer. In practice, we adopt a more direct way by normalizing the cluster centers $\bOmega_j, j\in[1,k]$ to have a length of $1$ (\textit{i.e.}, $\bOmega_j=\bOmega_j/\|\bOmega_j\|$). Accordingly, to preserve the validity of Euclidean distance, data points are normalized to have a unit length as well (\textit{i.e.}, $\beta = 1$). In this sense, $\bW_j$ would have a length of 2 and $\bb_j$ becomes a constant. As a result, the loss function in Eq.~(\ref{eq3.7}) could be rewritten into
\begin{equation}
	\cL = \sum_j \cL_j= \sum_{j}\cI_{j}(2-\bW_{j}^{\top}\bx).
\end{equation}

Note that since $\bW_j$ is optimized through SGD, in practice, we have to renormalize it after each update. However, as illustrated in Fig.~\ref{fig:normalization}(b), when the gradient is much larger than the length of $\bW_j$, $\bW_j$ will be greatly changed after each update. Taking the MNIST data set as an example, $\bW_{old}$ may correspond to the cluster center of digit ``3'' at first. However, when the gradient is considerably large, $\bW_{new}$ would shift to the center of digit ``5'' after optimization. In other words, the intrinsic semantic meaning of $\bW_j$ may differ in each iteration, which would cause unstable optimization, and thus the network is hard to converge.

Considering the aforementioned drawbacks, we propose simultaneously normalizing the weight and gradient as illustrated in Fig.~\ref{fig:normalization}(c). When the gradient is small enough, the cluster centers are mildly optimized and their semantic meanings keep the same across the training process, which promises a stable convergence. The ablation studies in Section~\ref{sec:ablation} proves the effectiveness of such a gradient normalization strategy. In practice, we experimentally normalize the gradient to 10\% of the length of $\bW_j$.

\subsubsection{End-to-end Training for Clustering and Representation Learning}
\label{sec:3.2.4}

\begin{figure}[!t]
\centering
\includegraphics[width=0.7\textwidth]{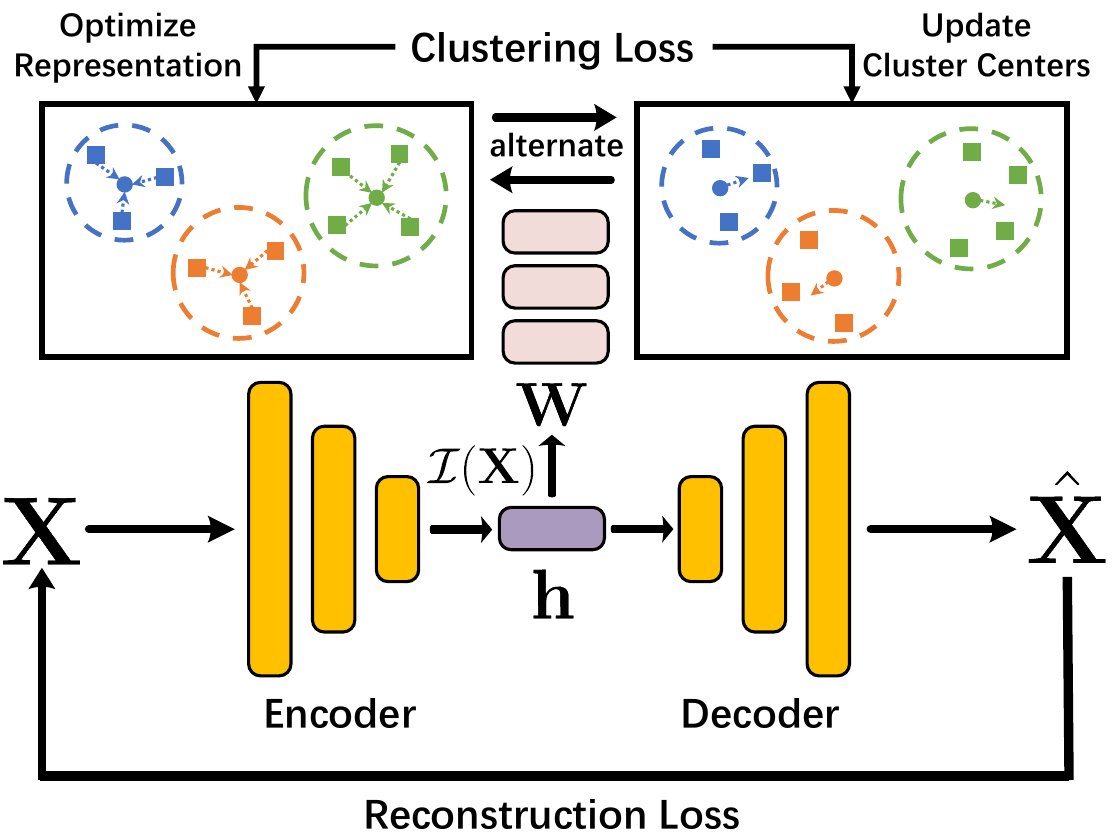}
\caption{\label{fig:end2end} The end-to-end training framework. The reconstruction loss is used to optimize encoder and decoder simultaneously, while the clustering loss is used to optimize the cluster layer and encoder alternatively.
}
\end{figure}

Based on the above discussions, we have recast the vanilla $k$-means to a neural layer with the following differentiable loss, namely,
\begin{equation}
\label{eq:clu_loss}
	\cL_{clu} = \sum_{i,j}\cI_{j}(\bX_{i})(2-\bW_{j}^{\top}\bX_i).
\end{equation}

Comparing with the vanilla $k$-means, one major advantage of our TELL is its plug-and-play characteristic, namely, it could be plugged into any neural network so that the deep representation could be utilized to boost the clustering performance. To this end, instead of directly conducting clustering in the raw feature space, we use an autoencoder (AE) to extracted more discriminative features $\bh=\{\bh_1, \bh_2, \dots\}$ by minimizing the following reconstruction loss, namely,
\begin{equation}
\centering
\label{eq:rec_loss}
\begin{aligned}
	\bh_i  = f(\bX_i), ~~~~~~~~~~\\
	\hat{\bX}_i = g(\bh_i),~~~~~~~~~~ \\
	\cL_{rec} = \sum_i \|\bX_i-\hat{\bX}_i\|_2^2,
\end{aligned}
\end{equation}
where $f(\cdot)$ and $g(\cdot)$ denote the encoder and decoder respectively, and $\bh_i$ is normalized to have a unit length as aforementioned. By replacing $\bX_i$ with $\bh_i$ in Eq.~(\ref{eq:clu_loss}), the overall loss of TELL is a combination of the reconstruction loss and the clustering loss, \textit{i.e.},
\begin{equation}
\label{eq:overall_loss}
\begin{aligned}
	\cL &= \cL_{rec} + \lambda \cL_{clu}\\
	&= \sum_i \|\bX_i-g(f(\bX_i))\|_2^2 + \lambda \sum_{i,j}\cI_{j}(\bX_{i})(2-\bW_{j}^{\top}f(\bX_i)),
\end{aligned}
\end{equation}
where $\lambda=0.01$ weights the two losses.

As can be seen, the reconstruction loss is used to simultaneously optimize the encoder $f(\cdot)$ and decoder $g(\cdot)$. For the clustering loss, we have shown that it can optimize the cluster layer weight $\bW_{j}$. Here, to further improve the representability of features, we also optimize the encoder $f(\cdot)$ with the clustering loss by pulling features to their corresponding cluster centers. In practice, to stabilize the training, the right term of Eq.~(\ref{eq:overall_loss}) is used to optimize $\bW$ and $f(\cdot)$ alternatively. The overall end-to-end training framework is summarized in Fig.~\ref{fig:end2end}.

\subsection{Interpretability of TELL}
\label{sec:3.3}

\begin{figure}[!t]
\centering
\includegraphics[width=0.8\textwidth]{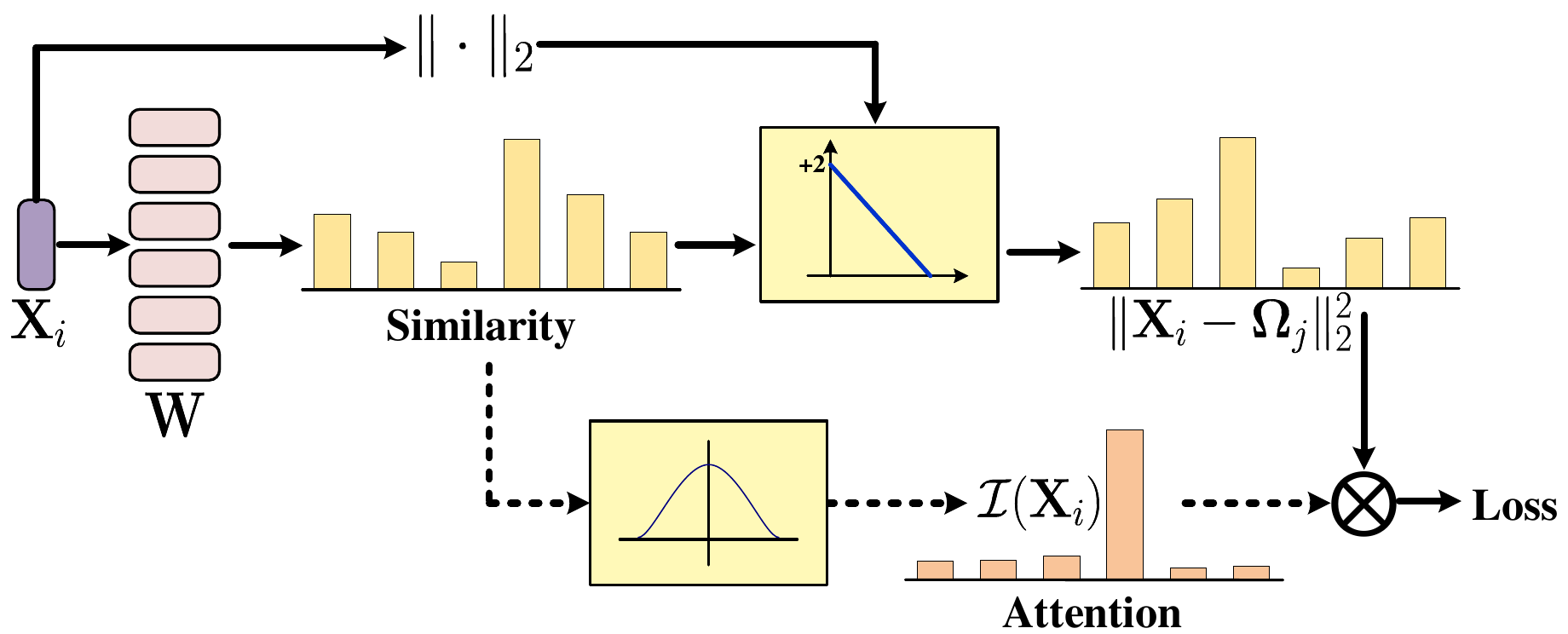}
\caption{\label{fig:attention} Besides the inhere interpretability, our TELL could also be understood from the perspective of the attention mechanism. The pathway below (denoted by the dot line) implements attention on the loss of the hyperplane $\bW$ \textit{w.r.t.} the input $\bX_i$. In brief, our TELL learns the attention $\bX_i$ paid on the cluster centers $\bOmega$.
}
\end{figure}

Although explainable artificial intelligence (XAI) has achieved remarkable progress recently~\citep{XAI}, one barrier to the consensus on common grounds is the interchangeable misuse of ``explainability'' and ``interpretability'' in the literature. In brief, explainability often refers to post-hoc explanations by various approaches to enhance the understandability of the model, such as text explanations, visual explanations, explanations by simplification, and feature relevance explanations techniques. Different from explainability, interpretability is rooted in the design of the model itself, which is highly expected but also quite challenging. The interpretability is also expressed as transparency, which includes the model decomposability and algorithmic transparency~\citep{XAI}. In the following, we will show that the proposed method (the cluster layer to be specific) enjoys these two interpretable characteristics.

Our TELL embraces model decomposability which stands for the feasibility to explain each part of the cluster layer. In other words, the input, weight parameters, activation, and loss function of our cluster layer are all interpretable. To be specific, the input to the cluster layer corresponds to the given data points, the weight $\bW$ is exactly the cluster centers $\bOmega$, the \textit{argmax} activation is used to achieve clustering by assigning each data point to its nearest cluster, and the loss function is recast from the vanilla $k$-means as shown in Eq.~(\ref{eq.3.4b}-\ref{eq.3.4c}). To strengthen our claim on the interpretability, we also made some post-hoc explanations by visualizing the learned cluster centers reconstructed from $\bOmega$. As shown in Fig.~\ref{fig:rec}, the reconstruction of cluster centers exactly corresponds to the MNIST digits, which demonstrates that TELL indeed captures the intrinsic semantic meanings.

Our TELL also possesses algorithmic transparency since its error surface or dynamic behavior can be reasoned about mathematically, allowing the user to understand how the model acts. To be specific, we not only theoretically provide the convergence analysis of our method later in Section~\ref{sec:3.4}, but also show the necessity of decoupling $\bW$ and $\bb$, as well as normalizing both the weight $\bW$ and its gradient, to achieve proper and stable optimization.

In addition, we could also understand the working manner of TELL from the standpoint of attention mechanism that is popular in natural language processing~\citep{Bahdanau:2014vz, Vaswani:2017ul}. As shown in Fig.~\ref{fig:attention}, TELL aims to learn a linear hyperplane  $\bW$ (denoted by the upper pathway) spanned by a set of cluster centers. The hyperplane is able to partition similar data points into the same cluster and dissimilar ones into different clusters based on attention. More specifically, to learn the hyperplane through Eq.~(\ref{eq:3.4}), TELL computes the dissimilarity between input $\bX_{i}$ and cluster centers $\bOmega$ via $(2-\bW^{\top}\bX_{i})$. After that, the loss of TELL is the summation of the weighted dissimilarity based on $\cI(\bX_{i})$. Intuitively, this implements the attention mechanism as shown in the pathway below in Fig.~\ref{fig:attention}, which decides the cluster centers $\bX_{i}$ pays attention to. Actually, the attention here serves as the clustering assignment.

\subsection{Convergence Proofs}
\label{sec:3.4}

In this section, we theoretically prove that the proposed loss $\cL$ sufficiently converges with the SGD optimization. Due to the space limitation, we provide full details of the proof in Appendix A, including some supporting experimental studies.

For ease of presentation, let $\mathcal{L}^{\ast}$ denote the optimal loss, $\mathcal{L}^{\ast}_{t}$ be the smallest loss so far at step $t$, and $\bW^{\ast}$ be the desirable weight which corresponds to the optimal cluster centers $\bOmega^{\ast}$. We consider the case that the standard SGD is used to optimize our network, \textit{i.e.},
\begin{equation}
	\label{eq:3.10}
	\bW_{t+1}=\bW_{t}-\eta_{t}\nabla\mathcal{L}(\bW_{t}),
\end{equation}
where $\nabla\mathcal{L}(\bW_{t})$ denotes the gradient of $\mathcal{L}$ \textit{w.r.t.} $\bW_{t}$. In the following, we abbreviate $\nabla\mathcal{L}(\bW_{t})$ to $\nabla\mathcal{L}_{t}$ for simplicity.  

\begin{ap_definition}[Lipschitz Continuity]
\label{def1}
	A function $f(x)$ is Lipschitz continuous on the set  $\Omega$, if there exists a constant $\epsilon>0$, $\forall x_{1}, x_{2}\in \Omega$ such that
	\begin{equation}
		\label{eq:3.11}
		\|f(x_{1})-f(x_{2})\| \leq \epsilon\|x_{1}-x_{2}\|,
	\end{equation}
where $\epsilon$ is termed as the Lipschitz constant.
\end{ap_definition}

Namely, the objective function $\mathcal{L}$ of TELL is Lipschitz continuous \textit{i.i.f.} $\|\nabla \mathcal{L}_{t}\|\leq \epsilon$. In other words, to meet the Lipschitz continuity, we need to prove that the upper boundary of $\nabla \mathcal{L}_{t}/\tau$ exists.

\begin{theorem}
	\label{thm1}
	There exists $\epsilon>0$ such that $\|\nabla \mathcal{L}_{t}\|\leq \epsilon$, where $\epsilon = \tau+2\tau\max(\|\bz_{i}\|)$ 
	and 
	$\bz_{i}=\bW_{i}^{\top}\bx/\tau$.
\end{theorem}

Theorem~\ref{thm1} shows that the proposed objective function $\cL(\bW)$ will be upper bounded by a positive real number $\epsilon$ when $\|\bz_{i}\|$ is bounded. As a matter of fact, there exists an upper boundary of $\|\bz_{i}\|$ for any real-world data set. Furthermore, without loss of generality, one could normalize $\bx$ and $\bOmega_i$ to meet $\|\bx\|=\|\bOmega_i\|=1$, and thus $\|\bW_i\|=2$ is bounded. Based on Theorem~\ref{lem1}, we have the following theorem.

\begin{theorem}
	\label{thm2}
	One could always find an optimal model $\cL_{T}^{\ast}$ which is sufficiently close to the optimal $\cL^{\ast}$ after $T$ steps, \textit{i.e.},
	\begin{equation}
		\label{thm2:eq1}
		\cL_{T}^{\ast}-\cL^{\ast}\leq \frac{\|\bW_{1}-\bW^{\ast}\|_{F}^{2}+\epsilon^{2}\sum_{t}^{T}\eta_{t}^{2}}{2\sum_{t=1}^{T}\eta_{t}}.
	\end{equation}
\end{theorem}

Based on Theorem~\ref{thm2}, we could derive the following two lemmas.
\begin{ap_lemma}
	\label{lem1}
	For the fixed step size (\textit{i.e.} $\eta_{t}=\eta$) and $T\rightarrow\infty$, 
\begin{equation}
	\label{lem1:eq1}
	\mathcal{L}^{\ast}_{T}-\mathcal{L}^{\ast}\rightarrow\frac{\eta \epsilon^{2}}{2}.
\end{equation}
\end{ap_lemma}

\begin{ap_lemma}
	\label{lem2}
	For the fixed step length (\textit{i.e.} $\eta_{t}=\eta/\nabla\mathcal{L}_{t}$) and $T\rightarrow\infty$, 
\begin{equation}
\label{lem2:eq1}
	\mathcal{L}^{\ast}_{T}-\mathcal{L}^{\ast}\rightarrow\frac{\eta \epsilon}{2}.
\end{equation} 
\end{ap_lemma}

Lemma~\ref{lem1} and \ref{lem2} show that the loss will eventually converge to $\mathcal{L}^{\ast}$ with a radius of $\frac{\eta\epsilon^{2}}{2}$ and $\frac{\eta \epsilon}{2}$ within $T$ steps.

\section{Related Works}
\label{sec2}

This work is closely related to XAI, clustering, and differentiable programming, which will be briefly introduced in this section.

\subsection{Model Explainability vs. Interpretable Model}

Generic deep architectures, as often referred to as ``black-box'' methods, rely on stacking somewhat ad-hoc modules, which makes it prohibitive to understand their working mechanisms. Despite a few hypotheses and intuitions, it appears difficult to understand why deep models work, how to analyze them, and how they are related to classical machine learning models. 

To solve the aforementioned problem, a variety of works~\citep{Zeiler:2014fr, Koh:2017ui, Bau:2017jj, Dosovitskiy:2016el, Kim:2016ts} have devoted towards the explainability of neural networks. In general, these works mainly focus on establishing some post-hoc explanations by designing some visualization techniques or agent models to enhance the understandability of neural networks.

Different from these studies on the explainability of neural networks, we directly develop a novel interpretable neural network as advocated in~\citep{Rudin:2019ha}. To be specific, the proposed TELL enjoys not only the post-hoc explainability but also the interpretability by design (see Section~\ref{sec:3.3}). In short, one could explicitly understand why the structure of the TELL is presented as itself, the physical meaning of each part of the cluster layer, and why it can perform data clustering.

\subsection{Stochastic $k$-means Clustering}

To enhance the scalability of the vanilla $k$-means, the stochastic approximation was first presented in~\citep{NIPS1994_989}, which is also called online $k$-means. Another pathway is generalizing the idea to mini-batch $k$-means~\citep{NIPS2016_6481, AISTATS:2017}. The major difference between stochastic and mini-batch $k$-means is that the former updates all centers asynchronously whereas the latter updates cluster centers after each iteration. Another difference is that mini-batch $k$-means is provable to converge to a local optimum, whereas it is not easy to promise that stochastic $k$-means could monotonically approximate the $k$-means objective, or even its expectation.

Compared with the stochastic methods, our TELL enjoys both online and mini-batch characteristics brought by its neural network implementation. In addition, instead of computing cluster centers as the mean of samples, TELL optimizes the network weight $\bW$ to obtain the centers via $\bOmega=\frac{1}{2}\bW$. As far as we know, there has not been any attempt like TELL before to establish a differential neural network for the vanilla $k$-means or its stochastic variants. Another advantage of TELL is that it could be plugged into any neural network to perform clustering and help the network to learn a clustering-favorable representation in an end-to-end manner (see Section~\ref{sec:3.2.4}).

\subsection{Differentiable Programming}

Differentiable programming (DP, also called model-based optimization) is an emerging and impactful topic. It bridges classical machine learning models and deep neural networks by emphasizing problem-specific prior and interpretability. DP advocates building complicated end-to-end machine learning pipelines by assembling parameterized functional blocks, that are later jointly trained from examples, using some form of differential calculus---mostly stochastic gradient descent (SGD). It bears resemblances to building software, except that it is parameterized, automatically differentiated, and trainable/optimizable. 

To the best of our knowledge, Learned ISTA (LISTA)~\citep{Gregor:2010} could be the first well-known DP work in the area of deep learning, which unfolds the ISTA algorithm~\citep{Blumensath2008:ISTA}, a popular $\ell_1$-optimizer, as a simple RNN. In the unrolled RNN, the number of layers and the weight correspond to the iteration number and the dictionary, respectively. Inspired by the success of LISTA, numerous methods have been proposed to address a variety of problems, \textit{e.g.} image restoration~\citep{Chen2015:On}, audio processing~\citep{Sprechmann2015:DeepRank},  segmentation~\citep{Zheng2015:CRFRNN}, hashing~\citep{Liu2018:TNNLS}, and  clustering~\citep{Wang2015:Learning}. 

As discussed in the Introduction section, this work is remarkably different from most existing DP approaches in either network structure (FNN vs. RNN) and applications (clustering vs. optimization), which may serve as a novel angle to facilitate future DP works.

\section{Experimental Results}
\label{sec4}

In this section, we carry out experiments to verify the effectiveness of the proposed TELL comparing with 14 state-of-the-art clustering approaches. Due to the space limitation, we present additional theoretical and experimental analyses in the attached supplementary material.

\subsection{Experimental Settings}
\label{sec4.1}

All experiments are conducted on a Nvidia 2080Ti GPU with PyTorch 1.7.0 and CUDA 11.0. For all the compared baselines, we use the source code released by authors.

\textbf{Baselines:} We compare TELL with \textbf{i)} three popular subspace clustering approaches including Spectral Clustering (SC)~\citep{Ng2002}, LRR~\citep{Liu2013}, and LSR~\citep{Lu2012}; \textbf{ii)} two large-scale clustering methods including Scalable LRR (SLRR)~\citep{Peng2015SRSC} and Large-scale Spectral Clustering (LSC)~\citep{Cai2015:LSC}; \textbf{iii)} two matrix decomposition based methods and agglomerative clustering methods, \textit{i.e.} NMF~\citep{Cai2011} and Zeta function based Agglomerative Clustering (ZAC)~\citep{Zhao2009:Cyclizing}; and \textbf{iv)} two deep learning based clustering methods, \textit{i.e.}, Deep Embedding Clustering (DEC)~\citep{Xie2016:Un} and Variational Deep Embedding (VaDE)~\citep{jiang2016variational}. Moreover, we also use the vanilla $k$-means, GMM, and FCM~\citep{Bezdek1981:FCM} as  baselines. Notably, either of LSR and LSC has two variants, which are denoted by LSR1/LSR2 and LSC-R/LSC-K, respectively.

\textbf{Implementation Details:} We adopt a convolutional autoencoder to extract 10-dimensional features for all the data sets and then pass the feature into our cluster layer (see Fig.~\ref{fig:end2end}). More specifically, the encoder consists of four convolutional layers $conv(16, 3, 1, 1)$-$conv(32, 3, 2, 1)$-$conv(32, 3, 1, 1)$-$conv(16, 3, 2, 1)$ followed by a two-layer MLP $fc(256)$-$fc(10)$, where $conv(16, 3,$ $1, 1)$ denotes a convolutional layer with a channel number of 16, a kernel size of 3, a stride length of 1, and a padding size of 1, and $fc(256)$ denotes a fully connected layer with 256 neurons.  Batch normalizations are applied after each convolutional layer, and the \textit{ReLU} activation is used at the end of each layer except the last. The decoder is mirrored from the encoder, with the \textit{sigmoid} activation at the output layer. Both the autoencoder and the cluster layer are randomly initialized with Kaiming uniform~\citep{he2015delving}, which are then simultaneously trained for 3000 epochs with the default Adadelta~\citep{Adadelta} optimizer. Motivated by the vanilla $k$-means, in practice, we run TELL five times with different random initializations and obtain the final result by the run with the minimal clustering loss $\cL_{clu}$ in Eq.~(\ref{eq:clu_loss})(see Appendix B for more details). For fair comparisons, we have tuned hyper-parameters for compared methods following the parameter tuning strategies suggested in the original papers and report them with corresponding results. Note that there is no hyper-parameter in the proposed method and no laborious tuning is needed.

\textbf{Data sets:} Our method is evaluated on the following three data sets, namely, the full MNIST handwritten digital database~\citep{Dataset:MNIST}, the full CIFAR-10 image database~\citep{Dataset:cifar10}, and the full CIFAR-100 image database~\citep{Dataset:cifar10}. For CIFAR-100, we adopt its 20 super-classes as partitions. In other words, we conduct experiments on the 20 super-classes of CIFAR-100 and report the mean, median, and maximum of the performance over these subsets, respectively. We normalize the data to be in the range of $[0, 1]$ before feeding them into the network, and no more preprocessing is applied.

\textbf{Evaluation Metrics:} Three widely used metrics are used to evaluate the clustering performance, including Clustering Accuracy (ACC), Normalized Mutual Information (NMI), and Adjusted Rand Index (ARI). For these three metrics, a higher value indicates a better clustering performance.

\subsection{Experimental Comparisons}
\label{sec4.2}

In this section, we evaluate the performance of TELL on three image benchmarks, including MNIST, CIFAR-10, and CIFAR-100. The training and test split are merged in all our experiments. Specifically, the MNIST data set consists of 70,000 handwritten digits over 10 classes, and each grayscale image is of size $28\times 28$. The CIFAR-10 data set consists of 60,000 RGB images of size $32\times 32 \times 3$ from 10 classes. The CIFAR-100 data set contains 60,000 RGB images of size $32\times 32 \times 3$ from 100 fine-grained classes which belong to 20 coarse-grained superclasses. The CIFAR-10/100 data sets are quite challenging and have been less touched in prior clustering works.

The clustering results on MNIST and CIFAR-10 are shown in Table~\ref{tab:1}. As can be seen, the proposed TELL gives superior clustering results in all three metrics. For example, the proposed TELL outperforms VaDE, a Gaussian Mixture Model (GMM) based deep clustering method, by 4.57\% in the term of ARI on MNIST, which proves its effectiveness. Note that LRR and SLRR show inferior performance on these two data sets, which may attribute to that the data does not meet the low-rank assumption well. We would like to point out that the performance of TELL could be further improved when a more powerful representation learning method is adopted, as verified in Table~\ref{tab:4}.

\begin{table*}[!t]
\centering\small
\begin{tabular}{l | cccr | cccr}
\toprule
\multicolumn{1}{c|}{\multirow{2}{*}{Methods}} & \multicolumn{4}{c|}{MNIST} & \multicolumn{4}{c}{CIFAR-10} \\
\cline{2-9}
 & \multicolumn{1}{c}{ACC} & \multicolumn{1}{c}{NMI} &\multicolumn{1}{c}{ARI} & \multicolumn{1}{c|}{Parameter} & \multicolumn{1}{c}{ACC} & \multicolumn{1}{c}{NMI} & \multicolumn{1}{c}{ARI} & \multicolumn{1}{c}{Parameter}\\
\midrule
$k$-means & 78.32 & 77.75 & 70.53 & --- & 19.81 & 5.94 & 3.01 & ---\\
GMM & 80.83 & 84.40 & 76.84 & --- & 19.31 & 7.06 & 3.33 & --- \\
FCM & 21.56 & 12.39 & 5.10 & --- & 17.02 & 3.92 & 2.56 & ---\\
SC & 71.28 & 73.18 & 62.18 & 1 & 19.81 & 4.72 & 3.22 & 10\\
LRR & 21.07 & 10.43 & 10.03 & 10.01 & 13.07 & 0.43 & 0.03 & 0.01\\
LSR1 & 40.42 & 31.51 & 21.35 & 0.4 & 19.79 & 6.05 & 3.64 & 0.6\\
LSR2 & 41.43 & 30.03 & 20.00 & 0.1 & 19.08 & 6.37 & 3.16 & 0.5\\
SLRR & 21.75 & 7.57 & 5.55 & 2.1 & 13.09 & 1.31 & 0.94 & 0.1\\
LSC-R & 59.64 & 56.68 & 45.98 & 6 & 18.39 & 5.67 & 2.58 & 3\\
LSC-K & 72.07 & 69.88 & 60.81 & 6 & 19.29 & 6.34 & 3.89 & 3\\
NMF & 46.35 & 43.58 & 31.20 & 10 & 19.68 & 6.20 & 3.21 & 3\\
ZAC & 60.00 & 65.47 & 54.07 & 20 & 5.24 & 0.36 & 0.00 & 10\\
DEC & 83.65 & 73.60 & 70.10 & 10 & 18.09 & 4.56 & 2.47 & 80\\
VaDE & 92.36 & 86.58 & 85.09 & --- & 20.87 & 7.20 & 3.95 & ---\\
\hline
TELL & \textbf{95.16} & \textbf{88.83} & \textbf{89.66} & --- & \textbf{25.65} & \textbf{10.41} & \textbf{5.96} & ---\\
\bottomrule
\end{tabular}
\caption{Clustering performance on the MNIST and CIFAR-10 data set.}
\label{tab:1}
\end{table*}

For CIFAR-100, we conduct clustering on its 20 super-classes of which each contains 3000 images from 5 fine-grained classes. As shown in Tables~\ref{tab:2} and \ref{tab:3}, the proposed TELL shows encouraging performance, which is 1.72\% and 3.30\% higher than its best competitor in terms of mean ACC on the first and last 10 super-classes, respectively. Comparing with the recently proposed DEC and VaDE, our method earns a performance gain of 6.43\% and 5.87\% on the last 10 super-classes, respectively. The results demonstrate the effectiveness of TELL and the benefits brought by the end-to-end training paradigm. Note that only the ACC metric is reported here due to the space limitation.

\begin{table*}[!t]
\centering
\resizebox{\textwidth}{!}{%
\begin{tabular}{l| cccccccccc | ccc }
\toprule
\multicolumn{1}{c|}{Methods} & \multicolumn{1}{c}{S1} &  \multicolumn{1}{c}{S2} & \multicolumn{1}{c}{S3} & \multicolumn{1}{c}{S4} & \multicolumn{1}{c}{S5} & \multicolumn{1}{c}{S6} & \multicolumn{1}{c}{S7} & \multicolumn{1}{c}{S8} & \multicolumn{1}{c}{S9} & \multicolumn{1}{c|}{S10} &  \multicolumn{1}{c}{Max} & \multicolumn{1}{c}{Mean} & \multicolumn{1}{c}{Median}\\
\midrule
$k$-means & 29.63 & 43.30 & 31.53 & 30.03 & 34.83 & 30.43 & 33.60 & \textbf{38.80} & 28.93 & 30.70 & 43.30 & 33.18 & 31.12\\
GMM & 28.37 & 38.07 & 28.80 & 27.53 & 32.10 & \textbf{34.10} & 32.30 & 33.57 & 29.67 & 28.43 & 38.07 & 31.29 & 30.89 \\
FCM & 26.77 & 37.80 & 25.30 & 25.97 & 29.77 & 26.37 & 32.60 & 36.73 & 25.00 & 25.33 & 37.80 & 29.16 & 26.57\\
SC & 31.90 & 39.30 & 33.67 & 27.53 & 34.27 & 27.77 & 33.10 & 36.17 & 26.90 & 32.30 & 39.30 & 32.29 & 32.70\\
LRR & 21.77 & 21.73 & 21.37 & 20.13 & 21.60 & 21.80 & 21.53 & 21.27 & 21.90 & 21.50 & 21.90 & 21.46 & 21.57\\
LSR1 & 21.93 & 21.40 & 22.27 & 21.87 & 21.47 & 21.30 & 22.33 & 21.97 & 21.07 & 21.90 & 22.33 & 21.75 & 21.89\\
LSR2 & 22.93 & 22.67 & 22.87 & 23.80 & 24.10 & 21.83 & 22.07 & 25.30 & 21.77 & 22.10 & 25.30 & 22.94 & 22.77\\
SLRR & 22.40 & 22.27 & 21.77 & 21.73 & 22.50 & 22.63 & 22.53 & 22.57 & 22.40 & 22.50 & 22.63 & 22.33 & 22.45\\
LSC-R & 31.97 & 40.50 & 30.77 & 28.87 & 34.30 & 28.67 & 32.90 & 35.27 & 27.13 & 32.03 & 40.50 & 32.24 & 32.00\\
LSC-K & 32.36 & 39.97 & \textbf{34.30} & 30.93 & 34.37 & 30.07 & 32.80 & 37.87 & 28.23 & 32.60 & 39.97 & 33.35 & 32.70\\
NMF-LP & 31.30 & 43.93 & 33.40 & 30.57 & 34.87 & 30.93 & 31.03 & 34.33 & 29.47 & 32.23 & 43.93 & 33.21 & 31.77\\
ZAC & 20.13 & 20.33 & 20.20 & 20.27 & 20.40 & 20.23 & 20.30 & 20.33 & 20.43 & 20.20 & 20.43 & 20.28 & 20.29\\
DEC & 31.17 & 43.97 & 29.97 & 30.60 & 34.87 & 28.50 & 33.40 & 20.07 & \textbf{29.87} & 31.97 & 43.97 & 31.44 & 30.89\\
VaDE & 28.47 & 35.83 & 23.83 & 25.67 & 35.23 & 29.57 & 33.10 & 36.53 & 28.20 & 26.27 & 36.53 & 30.27 & 29.02 \\
\hline
TELL & \textbf{34.07} & \textbf{46.20} & 30.03 & \textbf{31.47} & \textbf{37.30} & 31.03 & \textbf{36.90} & 38.73 & 29.83 & \textbf{35.13} & \textbf{46.20} & \textbf{35.07} & \textbf{34.60} \\
\bottomrule
\end{tabular}
\label{tab:2}
}
\caption{Clustering accuracy on the first 10 super-classes of the CIFAR-100 data set. }
\end{table*}

\begin{table*}[!t]
\centering
\resizebox{\textwidth}{!}{%
\begin{tabular}{l| cccccccccc | ccc }
\toprule
\multicolumn{1}{c|}{Methods} & \multicolumn{1}{c}{S11} &  \multicolumn{1}{c}{S12} & \multicolumn{1}{c}{S13} & \multicolumn{1}{c}{S14} & \multicolumn{1}{c}{S15} & \multicolumn{1}{c}{S16} & \multicolumn{1}{c}{S17} & \multicolumn{1}{c}{S18} & \multicolumn{1}{c}{S19} & \multicolumn{1}{c|}{S20} &  \multicolumn{1}{c}{Max} & \multicolumn{1}{c}{Mean} & \multicolumn{1}{c}{Median}\\
\midrule
$k$-means & 39.53 & 28.37 & 25.23 & 26.87 & 24.10 & 31.83 & 27.50 & 32.83 & \textbf{31.00} & 39.80 & 39.80 & 30.71 & 29.69\\
GMM & 30.10 & 30.03 & 27.07 & \textbf{29.50} &  24.83 & 32.80 & 27.70 & 31.83 & 29.27 & 38.83 & 38.83 & 30.20 & 29.77 \\
FCM & 41.80 & 29.33 & 23.77 & 26.77 & 23.30 & 29.40 & 27.43 & 23.27 & 25.70 & 32.97 & 41.80 & 28.37 & 27.10\\
SC & 40.77 & 31.80 & 24.83 & 26.33 & 23.97 & 31.03 & 30.57 & 30.97 & 28.50 & 39.30 & 40.77 & 30.81 & 30.77\\
LRR & 21.87 & 21.67 & 21.97 & 21.67 & 20.37 & 21.27 & 21.77 & 22.00 & 21.20 & 21.47 & 22.00 & 21.53 & 21.67\\
LSR1 & 21.80 & 21.93 & 21.57 & 21.30 & 22.10 & 22.27 & 22.00 & 21.70 & 21.60 & 21.90 & 22.27 & 21.82 & 21.85\\
LSR2 & 26.43 & 22.13 & 20.30 & 24.10 & 22.00 & 21.97 & 21.47 & 21.07 & 21.17 & 24.60 & 26.43 & 22.52 & 21.99\\
SLRR & 22.63 & 22.50 & 22.03 & 22.90 & 22.27 & 21.57 & 21.47 & 22.77 & 23.30 & 22.37 & 23.30 & 22.38 & 22.44\\
LSC-R & 41.53 & 30.87 & 24.47 & 26.43 & 23.70 & 20.97 & 29.10 & 31.97 & 28.63 & 32.93 & 41.53 & 29.06 & 28.87\\
LSC-K & 43.90 & 30.67 & 24.67 & 26.57 & 24.10 & 29.10 & \textbf{30.77} & 30.37 & 29.57 & 38.60 & 43.90 & 30.83 & 29.97\\
NMF-LP & 42.00 & 30.27 & 25.00 & 25.33 & 22.83 & 30.33 & 29.13 & 32.13 & 29.13 & 40.97 & 42.00 & 30.71 & 29.70\\
ZAC & 20.20 & 20.23 & 20.30 & 20.27 & 20.23 & 20.30 & 20.30 & 20.27 & 20.23 & 20.23 & 20.30 & 20.26 & 20.25\\
DEC & 21.80 & 20.17 & 25.03 & 26.90 & 23.80 & 31.83 & 27.07 & 28.57 & 30.63 & 41.17 & 41.17 & 27.70 & 26.99\\
VaDE & 31.07 & 32.00 & 27.17 & 28.47 & \textbf{25.73} & 24.63 & 28.23 & 21.87 & 24.57 & 38.90 & 38.90 & 28.26 & 27.70 \\
\hline
TELL & \textbf{46.70} & \textbf{35.93} & \textbf{29.73} & 29.10 & 24.63 & \textbf{35.93} & 30.53 & \textbf{37.37} & 27.57 & \textbf{43.83} & \textbf{46.70} & \textbf{34.13} & \textbf{33.23} \\
\bottomrule
\end{tabular}
}
\caption{Clustering accuracy on the last 10 super-classes of the CIFAR-100 data set.}
\label{tab:3}
\end{table*}

\subsection{Visualization Analyses}

To give a more intuitive understanding of the proposed TELL, in this section, we conduct two visualization analyses, including cluster center reconstruction and t-SNE visualization.

\textbf{Cluster Center Reconstruction:} We feed the cluster representations into the decoder and reconstruct the clustering centers learned by TELL and AE+$k$-means across the training process. According to reconstruction outcomes in Fig.~\ref{fig:rec}, both TELL and $k$-means learn more representative cluster centers as the training goes. However, $k$-means could fail to capture the intrinsic 10 digit numbers and mix confusing digits like `3'-`8' or `3'-`5' up. Notably, TELL also confuses the digit `4' and `9' at the early training stage (see results at epoch 1000), but it successfully distinguishes them after more iterations. In other words, TELL is more likely to achieve the global optimum than the vanilla $k$-means.

\def \m_wid{0.325\textwidth}
\begin{figure*}[!th]
\begin{center}
\subfigure [TELL Epoch=100]{\label{fig5a}\includegraphics[width=\m_wid]{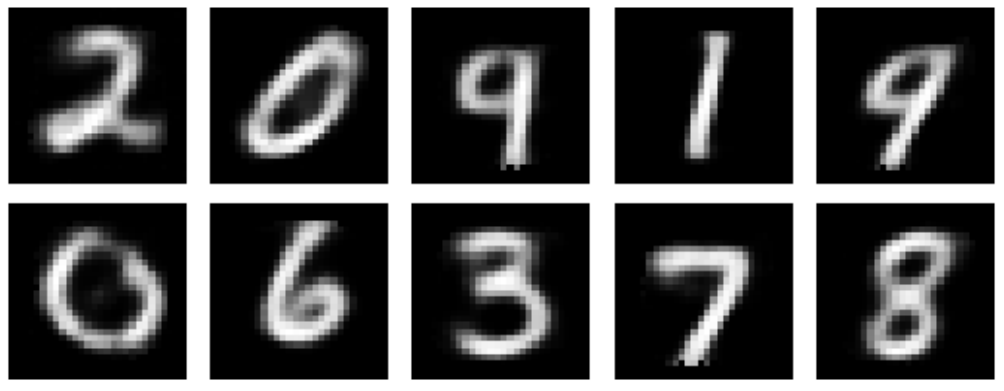}}~~~
\subfigure [TELL Epoch=1000]{\label{fig5b}\includegraphics[width=\m_wid]{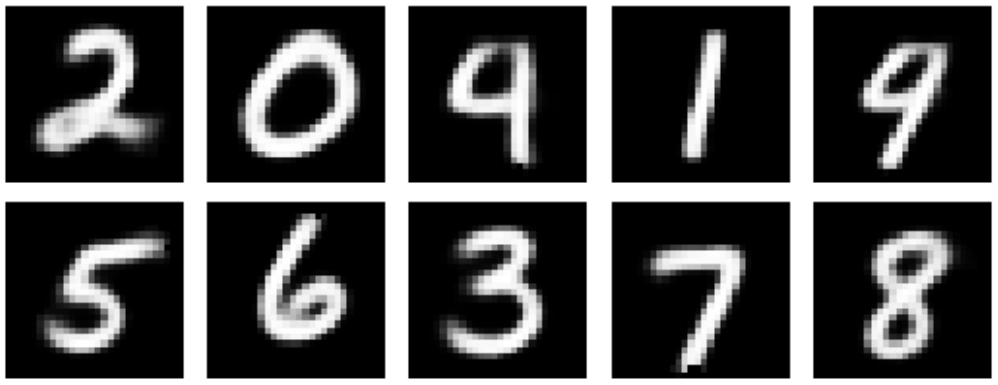}}~~~
\subfigure [TELL Epoch=3000]{\label{fig5c}\includegraphics[width=\m_wid]{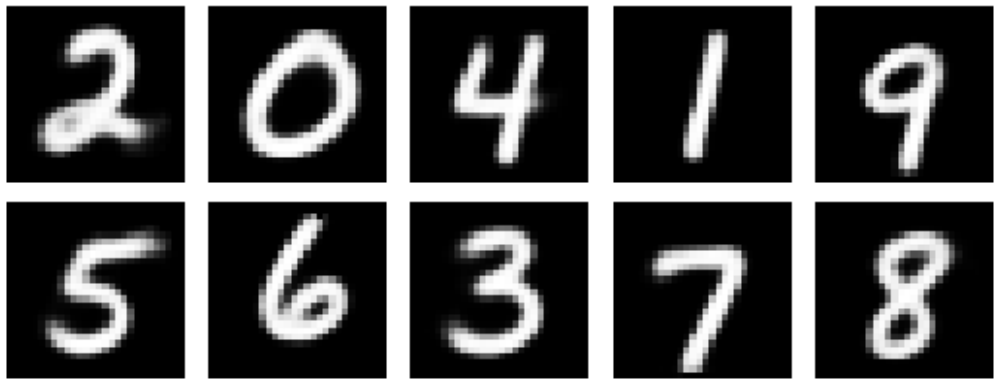}}
\subfigure [AE+$k$-means Epoch=100]{\label{fig5a}\includegraphics[width=\m_wid]{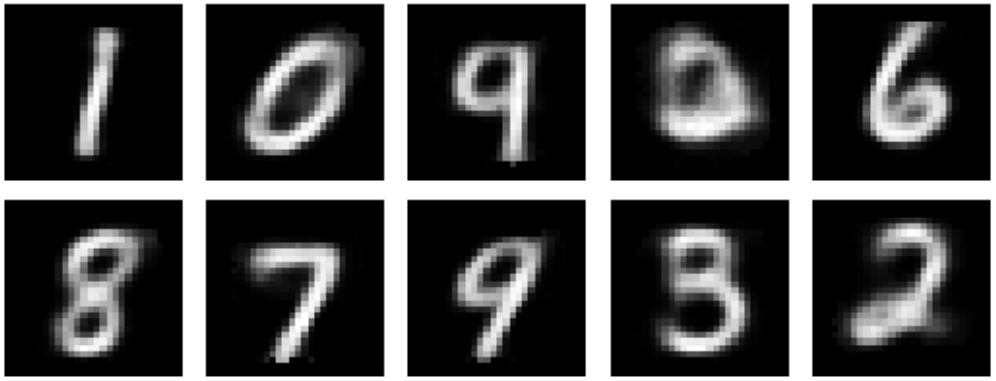}}~~~
\subfigure [AE+$k$-means Epoch=1000]{\label{fig5b}\includegraphics[width=\m_wid]{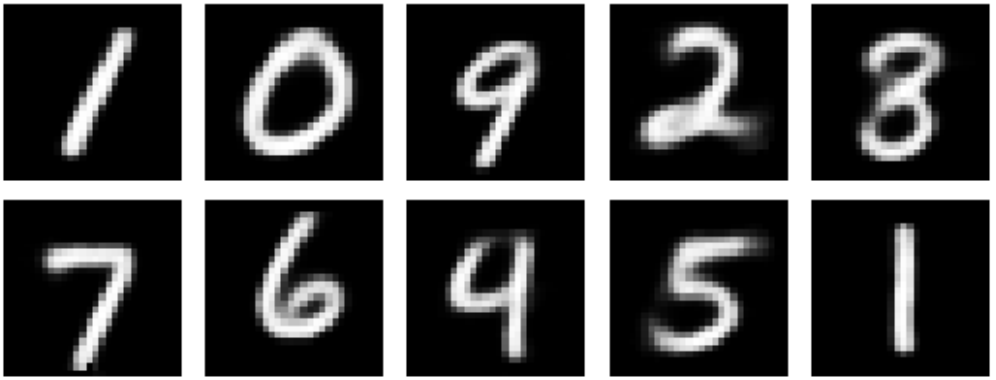}}~~~
\subfigure [AE+$k$-means Epoch=3000]{\label{fig5c}\includegraphics[width=\m_wid]{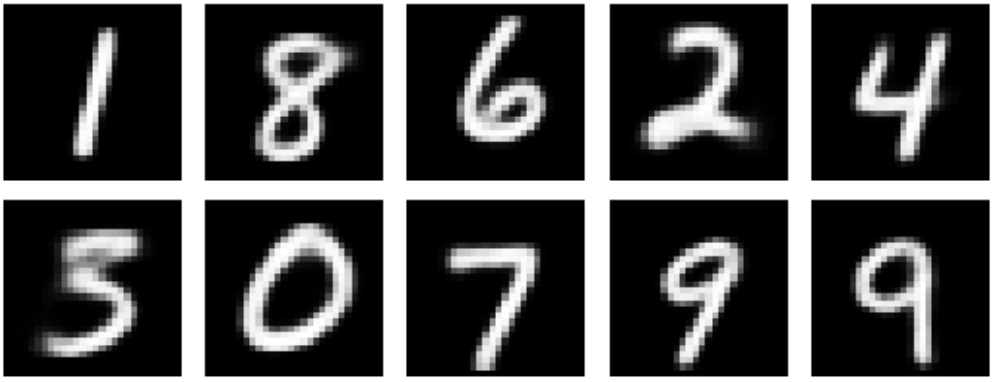}}
\end{center}
\caption{\label{fig:rec} Cluster centers reconstruction on the MNIST data set.}
\end{figure*}

\textbf{t-SNE Visualization:} To visualize the clustering result, we employ t-SNE~\citep{maaten2008visualizing} to reduce the dimensionality of the learned representation to two. As shown in Fig.~\ref{fig5a}, as the training goes, TELL learns a more compact and discriminative representation, which improves the separability of the estimated cluster centers.

\def \m_wid{0.325\textwidth}
\begin{figure*}[!t]
\begin{center}
\subfigure [epoch=0]{\label{fig5a}\includegraphics[width=\m_wid]{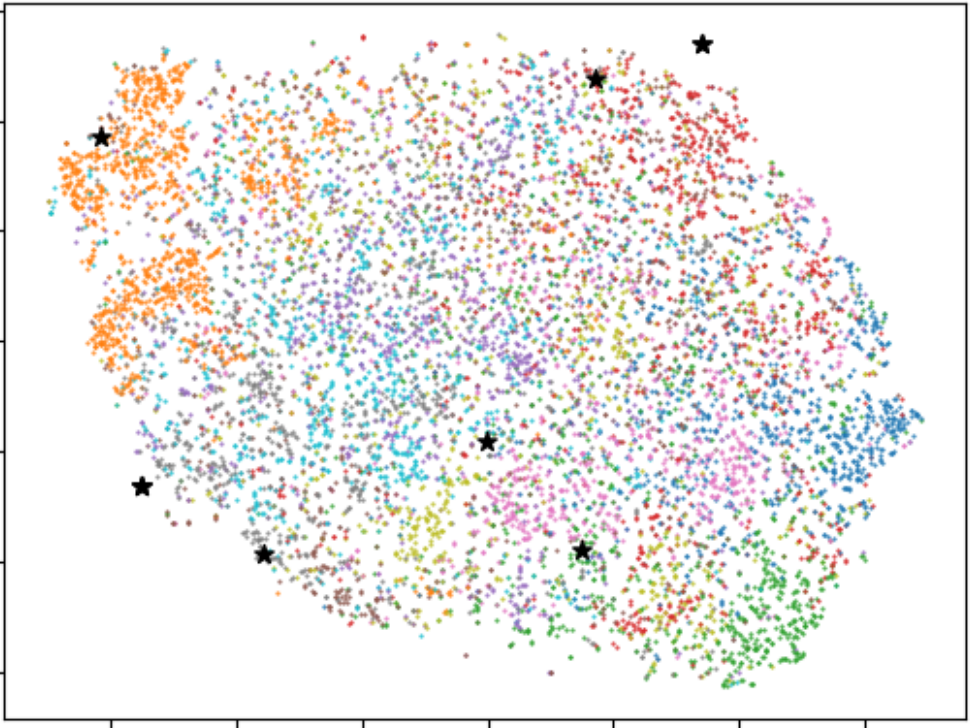}}
\subfigure [epoch=50]{\label{fig5b}\includegraphics[width=\m_wid]{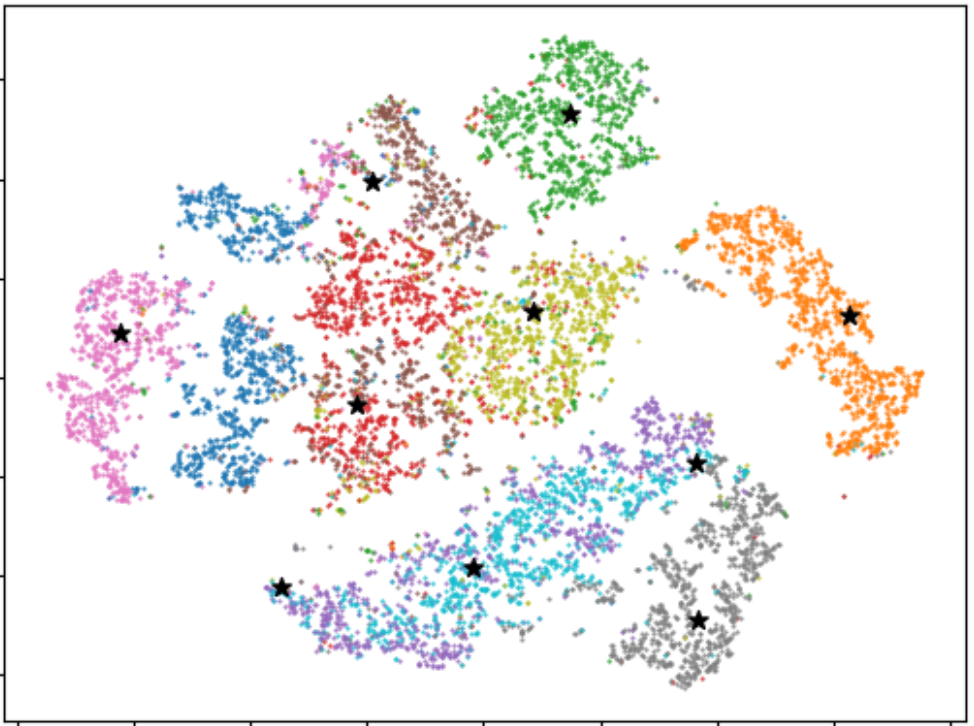}}
\subfigure [epoch=500]{\label{fig5c}\includegraphics[width=\m_wid]{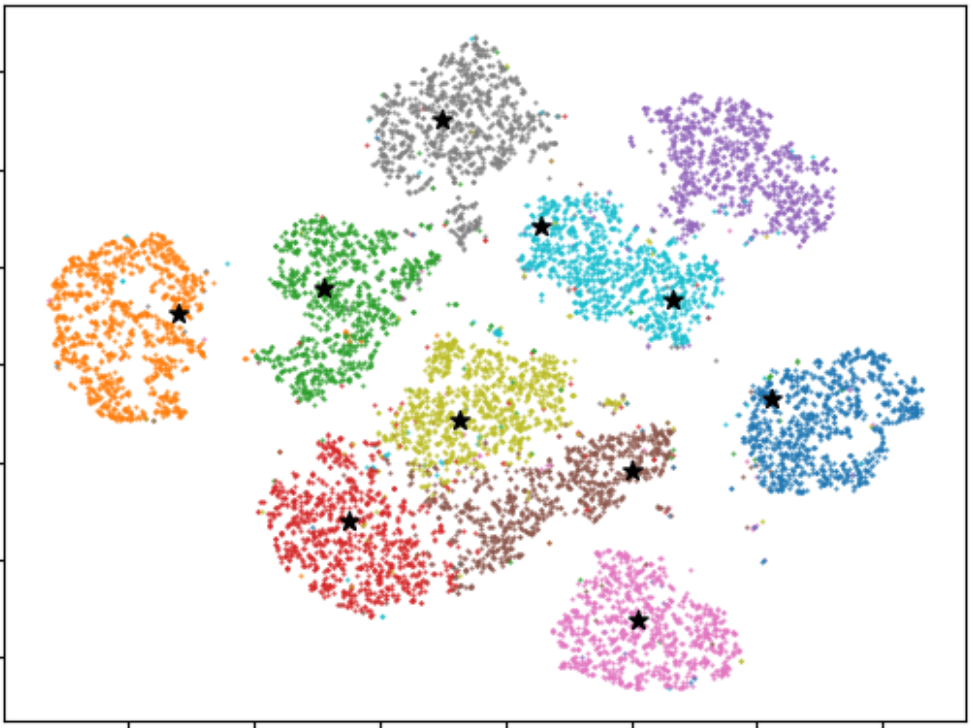}}
\subfigure [epoch=1000]{\label{fig5d}\includegraphics[width=\m_wid]{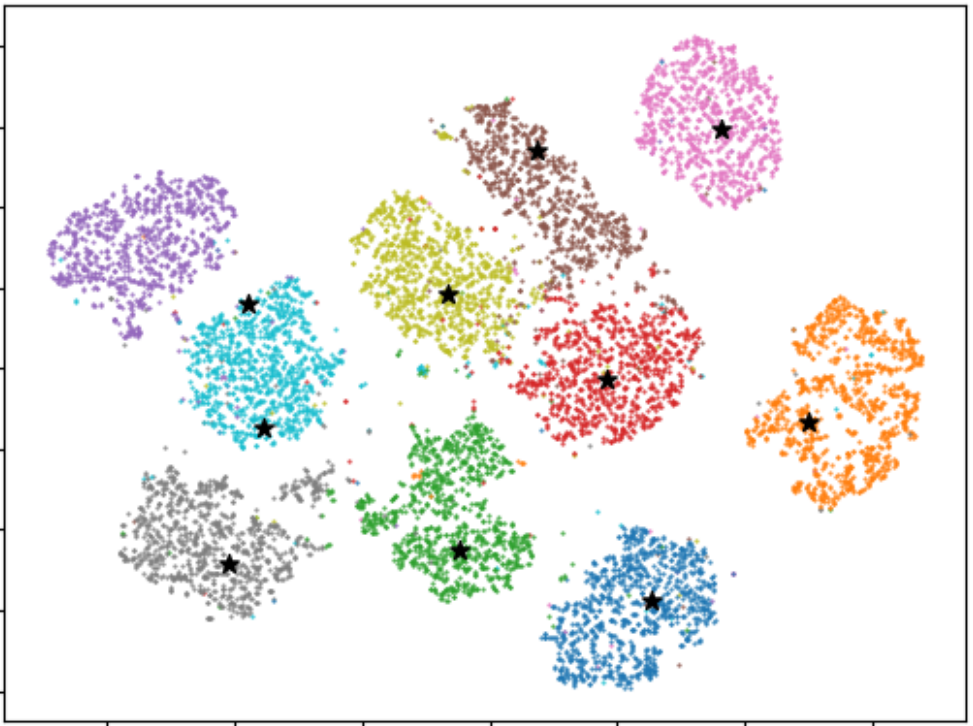}}
\subfigure [epoch=1500]{\label{fig5e}\includegraphics[width=\m_wid]{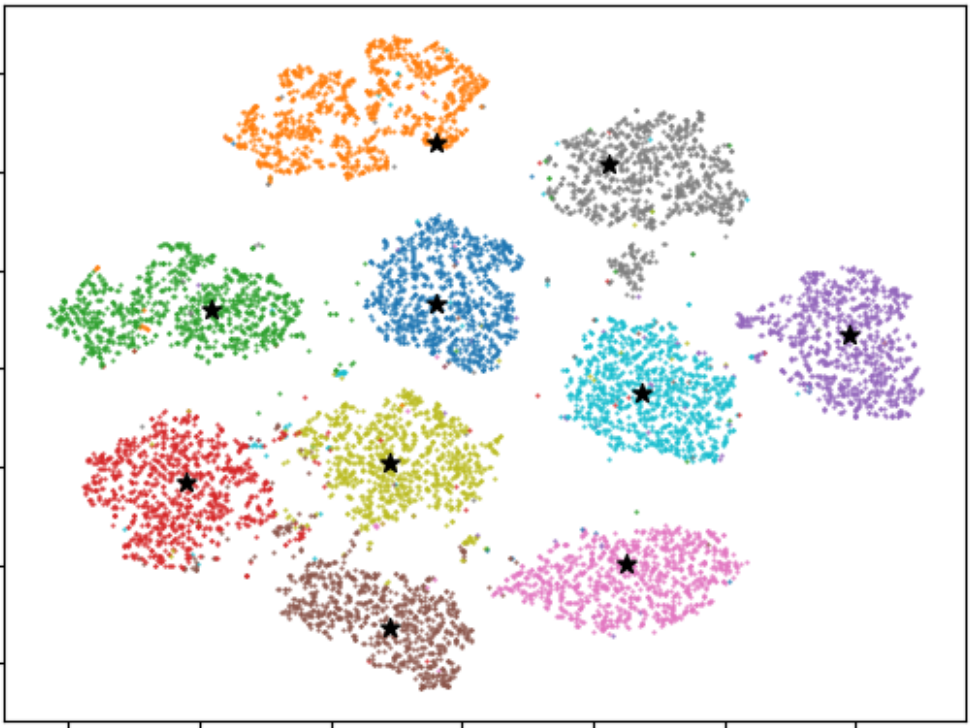}}
\subfigure [epoch=3000]{\label{fig5f}\includegraphics[width=\m_wid]{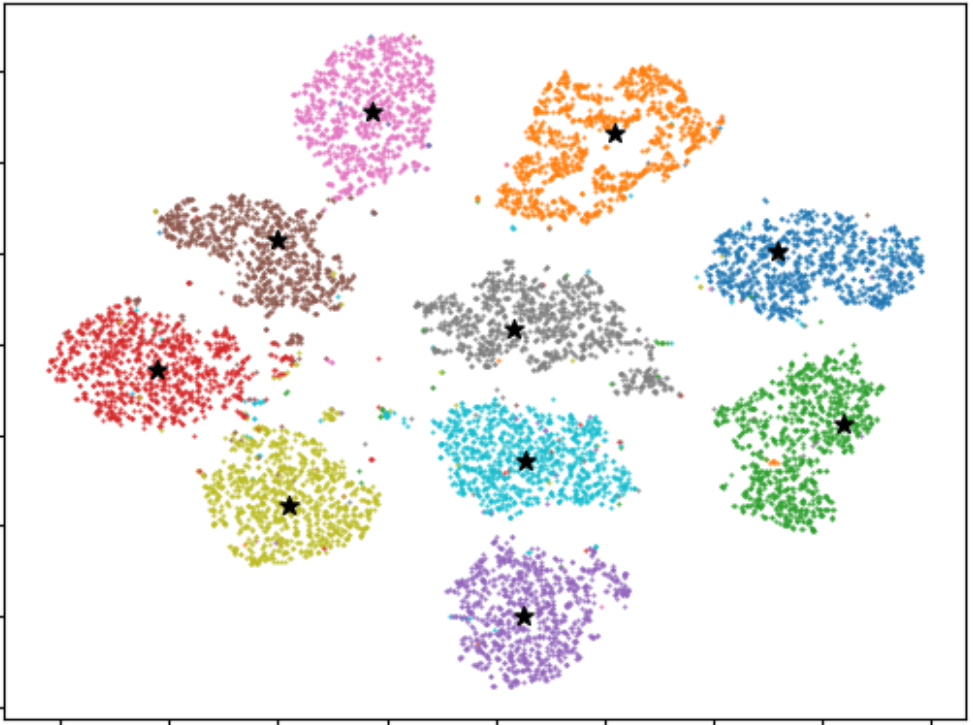}}
\end{center}
\caption{\label{fig5} t-SNE visualization on the learned MNIST representations across the training process. Different digits are denoted with different colors, and the black pentagrams denote the cluster centers estimated by TELL.}
\end{figure*}

\subsection{Ablation Studies}
\label{sec:ablation}
In this section, we conduct three ablation studies to investigate the robustness and effectiveness of the proposed TELL. Specifically, we test the performance of TELL with different training paradigms, features, and optimization strategies.

\textbf{Effectiveness of End-to-end Training:} In our method, the clustering loss is used to iteratively update cluster centers and optimize instance features. To prove the effectiveness of such an end-to-end learning paradigm, we test TELL on fixed features learned by autoencoders, \textit{i.e.}, the clustering loss is used to update cluster centers only.

In practice, we adopt a convolutional autoencoder to extract features from images with different channels (\textit{e.g.}, 1 for MNIST and 3 for CIFAR-10/100). Here, to see how our TELL relies on the feature extraction ability of the network, we also test TELL with the fully connected autoencoder on the MNIST data set. The structure of encoder is $fc(500)$-$fc(500)$-$fc(2000)$-$fc(10)$, and the decoder is symmetric. The \textit{ReLU} activation is applied at the end of each layer except the last layer of encoder where no activation is used and the last layer of decoder where \textit{sigmoid} is applied instead.

\begin{table*}[!t]
\centering
\begin{tabular}{c|ccc|ccc}
\toprule
\multirow{2}{*}{Method} & \multicolumn{3}{c|}{FCN} & \multicolumn{3}{c}{CNN}  \\ \cline{2-7} 
            & ACC   & NMI   & ARI   & ACC   & NMI   & ARI  \\ \hline
$k$-means & 78.32 &	77.75 &	70.53 & 80.40 & 79.65 & 74.34      \\ 
TELL-TwoStage          &   80.68    &  77.12     &    72.27   &   84.07    &    79.04   &   74.52       \\
TELL          & \textbf{93.25} & \textbf{85.52} & \textbf{85.76}   & \textbf{95.16} & \textbf{88.83} & \textbf{89.66}    \\
\bottomrule
\end{tabular}
\caption{Influence of different network structures on the MNIST data set. ``TwoStage'' means TELL is conducted on the fixed representations learned by the autoencoder instead of the standard end-to-end training.}
\label{tab:5}
\end{table*}

Table~\ref{tab:5} shows that the performance of TELL with end-to-end training is much better than the ``Two-Stage'' paradigm, which suggests that pulling features to its corresponding cluster center helps the network to learn clustering-favorable representations. It is also interesting to note that under the ``Two-Stage'' learning paradigm, TELL finds slightly better cluster centers with higher ACC and ARI but a bit lower NMI than the vanilla $k$-means. Besides, TELL achieves slightly inferior performance with FCN than CNN due to the weaker feature extraction ability. But we would like to point out that the clustering accuracy of TELL with FCN still outperforms all the 14 compared baselines.

\textbf{Influence of Feature Representability:} We notice that the deep clustering method IMSAT~\citep{Hu2017:ICML} could achieve a promising result by employing some data augmentation techniques. It achieves 98.4\% ACC on MNIST compared with 95.16\% by TELL. However, we would like to point out that TELL enjoys the following advantages. On the one hand, TELL is a transparent neural model which could be interpreted from perspectives of model decomposability, algorithmic transparency, and post-hoc explainability. On the other hand, TELL is a plug-and-play online clustering module complementary to any neural network. In other words, the clustering performance of TELL would benefit from a better representation learning module. To see how the proposed TELL relies on the quality of representations, we test TELL on the raw data and features learned by IIC~\citep{ji2019invariant}, which uses data augmentations to enhance the feature extraction ability. As shown in Table~\ref{tab:4}, TELL achieves a higher ACC than IMSAT. Besides, consistent with the results in Table~\ref{tab:5}, TELL could find slightly better cluster centers than the vanilla $k$-means with batch-wise optimization, while the latter needs the entire data set and is impractical for large-scale or online data sets.

\begin{table*}[!t]
\centering
\begin{tabular}{c|ccc|ccc}
\toprule
\multirow{2}{*}{Method}  & \multicolumn{3}{c}{Raw} & \multicolumn{3}{|c}{IIC} \\ \cline{2-7}
                                       & ACC & NMI & ARI & ACC & NMI & ARI \\ \hline
$k$-means      &  55.34   &  53.20   & 40.28    &  98.20   &  96.08   &  96.06   \\
TELL-TwoStage           &   56.20    &   52.82    &   40.33     & 98.64    &   96.54  &   96.98  \\ \hline
$k$-means$^\dagger$ &   48.52  &  46.75   &  31.25   &  98.10  &  95.90  &    95.85  \\
TELL-TwoStage$^\dagger$ &   54.59  &  50.80  &  38.33 &  98.44   &  96.22   &   96.56  \\
\bottomrule
\end{tabular}
\caption{Influence of features with different representability on the MNIST data set. `$\dagger$' indicates that the network is trained on 1,000 samples randomly selected from the test set, and evaluated on 60,000 training samples in an online manner.}
\label{tab:4}
\end{table*}

\textbf{Effectiveness on handling online data:} To see how well TELL handles new coming data, we also test it under a different online scenario wherein the model is first trained on a relatively small data set and then evaluated on a large online data set. Specifically, 1,000 digits are randomly sampled from the MNIST test set as the training data on which we apply the vanilla $k$-means and train our TELL to obtain the clustering centers. After that, 60,000 digits from the MNIST training set are used to evaluate the performance. During the evaluation, the data is presented in an online manner (\textit{i.e.}, each time only a batch of data comes, and it will not be accessible afterward). As the vanilla $k$-means could not update cluster centers without accessing the entire data set, clustering is achieved by assigning each digit to the closest cluster center computed on the early 1,000 digits. On the contrary, our TELL could timely update parameters of the cluster layer to fit new coming data through Eq.~(\ref{eq:clu_loss}). The last two rows in Table~\ref{tab:4} indicate that such a batch-wise optimization of cluster centers improves the clustering performance on online data, especially when there is a fair degree of biases between the training and test set. Note that here the superior performance of TELL than the vanilla $k$-means solely attributes to its timely update of cluster centers, since the feature is fixed and will not be optimized through end-to-end representation learning introduced in Sec~\ref{sec:3.2.4}.

\textbf{Influences of Optimizers:} Besides the above investigations on different initializations and features, we further consider the role of the used optimization strategies. To this end, we carry out experiments on the MNIST data set by training TELL with two popular SGD variants, namely, Adadelta~\citep{Adadelta} and Adam~\citep{Adam}. In the implementation, we adopt the default setting for these optimizers. For a more comprehensive study, we adopt four more metrics to evaluate the clustering quality, \textit{i.e.} Adjusted Mutual Index (AMI), Homogeneity (Homo.), Completeness (Comp.), and V\underline{~~}Measure (V\underline{~~}Mea.). Note that ACC, AMI, ARI, and AMI are external metrics that are computed based on the ground-truth, while Homogeneity, Completeness, and V\_Measure are internal metrics that measure the compactness/divergence of within-/between-cluster samples. Table~\ref{tab:6} shows that TELL with Adadelta performs slightly better than with Adam, and it is not necessary to tune the optimizer parameter.

\begin{table*}[!t]
\centering
\begin{tabular}{c| ccccccc }
\toprule
Optimizers & ACC & NMI & ARI & AMI & Homo. & Comp. & V\underline{~~}Mea.\\
\midrule
AdaDelta & 95.16 & 88.83 & 89.66 & 88.83 & 88.80 & 88.86 & 88.83 \\
Adam & 93.75 & 87.36 & 87.03 & 87.36 & 87.34 & 87.38 & 87.36 \\ \hline
AdaDelta w/o grad. norm. & 81.08 & 82.85 & 76.09 & 82.84 & 81.69 & 84.04 & 82.85 \\
Adam w/o grad. norm. & 80.49 & 80.48 & 73.36 & 80.47 & 79.32 & 81.67 & 80.48 \\
\bottomrule
\end{tabular}
\caption{Influence of different optimizers on the MNIST data set.}
\label{tab:6}
\end{table*}

Recall that to stabilize the update of cluster centers, we normalize their gradient to have an L2-norm of $0.1$, which is $10\%$ of the L2-norm of cluster centers. Here, to show the necessity of such a gradient normalization strategy, we conduct the following ablation study by simply applying the standard gradient descent on the cluster layer. One could see that both two optimizers give inferior performance when no normalization is performed. Because when the length of the gradient is much larger than that of cluster centers, the cluster centers change exceedingly at every iteration, thus preventing them from convergence.

\begin{figure*}[!t]
\begin{center}
\subfigure [TELL with FCN.]{\label{fig4a}\includegraphics[width=0.49\textwidth]{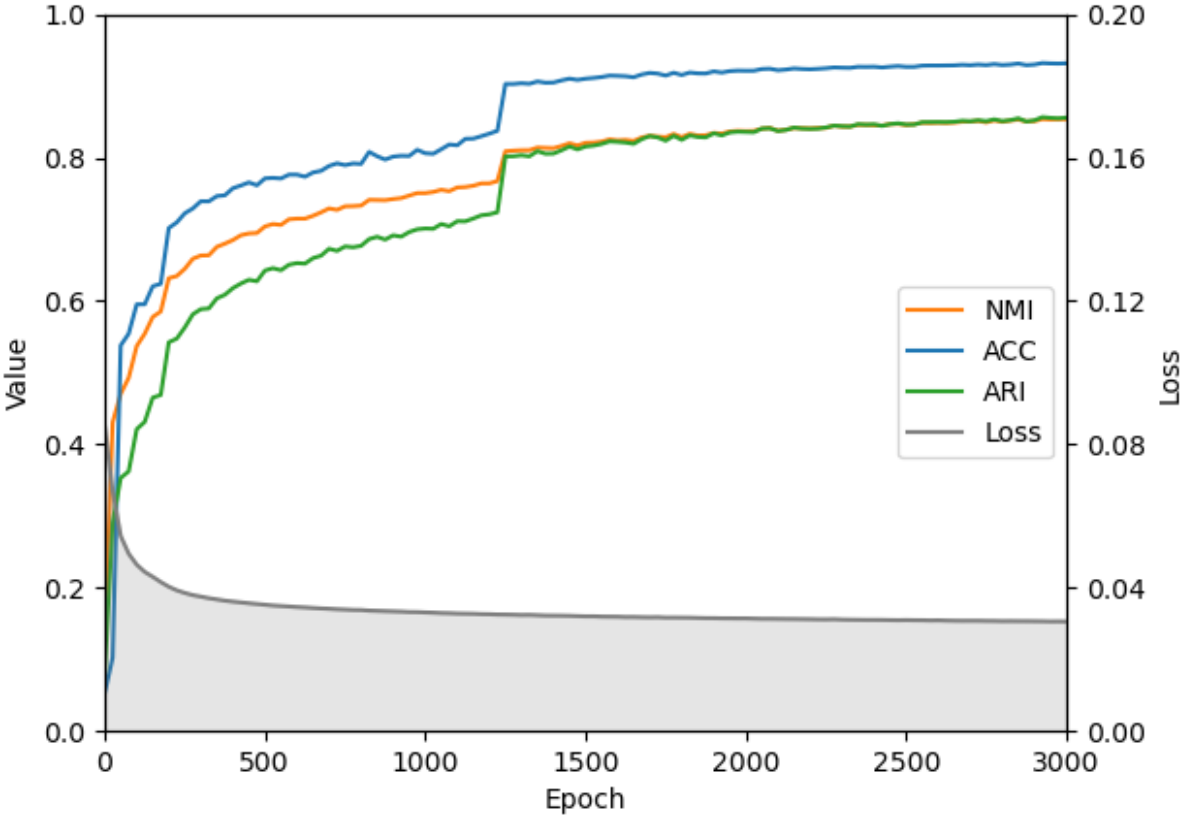}}
\subfigure [TELL with CNN.]{\label{fig4b}\includegraphics[width=0.49\textwidth]{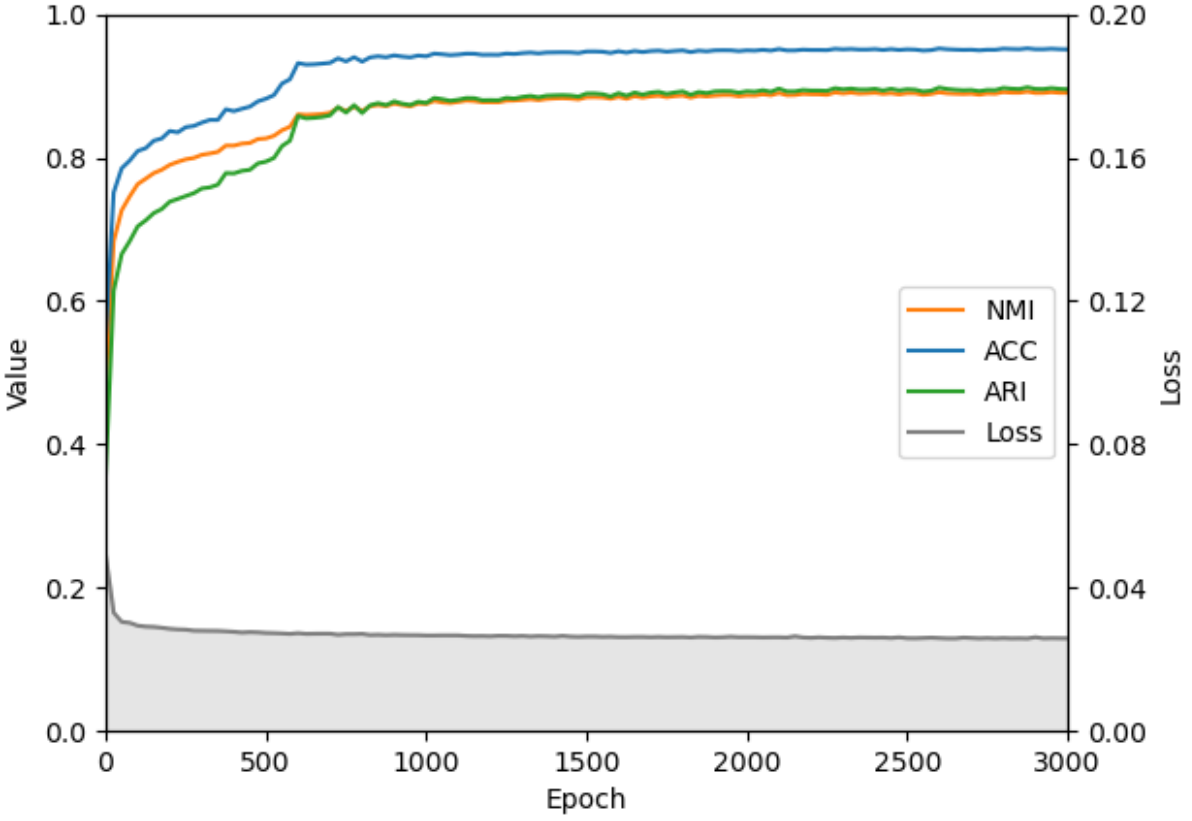}}
\end{center}
\caption{\label{fig4} Performance w.r.t training epoch on the MNIST data set. The left and right y-axis denote the clustering metrics and the loss, respectively.}
\end{figure*}

\subsection{Convergence Analysis}
In Section~3.2, we have theoretically proved that our method will sufficiently approximate the global optimum under some mild conditions. In this section, we conduct experiments on the MNIST data set to verify our theoretical analysis. In \figurename~\ref{fig4}, we report the clustering performance and the loss value of our method with the fully connected neural network and the convolutional neural network. From the result, one could observe that TELL convergences after $\scriptsize{\sim}$1400/800 epochs with FCN/CNN in terms of NMI, ACC, ARI, and the loss value. It should be pointed out that the results reported in Table~\ref{tab:1}---\ref{tab:3} are with 3000 training epochs, and if we continue training TELL to 5000 epochs, the performance could be further improved. Namely, TELL achieves better results of 93.62\%, 86.25\%, 86.47\% and 95.54\%, 89.72\%, 90.44\% in terms of ACC, NMI, and ARI with FCN and CNN, respectively.

\section{Conclusion}
\label{sec5}

In this paper, we directly build an interpretable neural layer which is a differentiable alternative of the vanilla $k$-means. The proposed clustering neural network overcomes some shortcomings of the vanilla $k$-means and owns the properties of parallel computing, provable convergence, online clustering, and clustering-favorable representation learning. In addition, our method enjoys interpretability in terms of model decomposability, algorithmic transparency, and post-hoc explainability. Such interpretability is inhered in the model itself rather than by an agent model, which is highly expected in XAI. It should be pointed out that this paper only focuses on the interpretability of the cluster layer. In the future, we plan to investigate how to create an interpretable neural module for representation learning as well.


\acks{The authors would thank to the handing editor Prof. Blei and anonymous reviewers for the constructive comments and valuable suggestions that greatly improve this work, and also thank Dr. Huan Li and Prof. Zhouchen Lin to provide helps and initial theoretical analysis on the decoupling of the network weight and bias. This work was supported in part by the National Key R\&D Program of China under Grant 2020YFB1406702, in part by NFSC under Grant U19A2081, 61625204, and 61836006; in part by the 111 Project under grant B21044; in part by ARC under Grant DP180100106 and DP200101328; in part by A*STAR AME Programmatic Funding Scheme A18A1b0045 and A18A2b0046.}


\newpage

\appendix
\section*{Appendix A.}
\label{app:theorem}

\setcounter{theorem}{0}
\setcounter{ap_definition}{0}
\setcounter{ap_lemma}{0}
\setcounter{equation}{0}
\setcounter{figure}{0}

In this appendix, we provide full details of the convergence analysis of our TELL. In brief, we theoretically prove that the loss $\cL$ of our network will sufficiently converge with the SGD optimization.

For ease of presentation, let $\mathcal{L}^{\ast}$ denote the optimal loss, and $\mathcal{L}^{\ast}_{t}$ be the smallest loss found so far at the $t$-th step. Similarly, $\bW^{\ast}$ denotes the optimal weight corresponds to cluster centers $\bOmega^{\ast}$. We consider the case when the standard SGD is used to optimize our network, \textit{i.e.}
\begin{equation}
	\label{eq:3.8}
	\bW_{t+1}=\bW_{t}-\eta_{t}\nabla\mathcal{L}(\bW_{t}),
\end{equation}
where $\nabla\mathcal{L}(\bW_{t})$ denotes the gradient of $\mathcal{L}$ \textit{w.r.t.} $\bW_{t}$. In the following, we abbreviate $\nabla\mathcal{L}(\bW_{t})$ to $\nabla\mathcal{L}_{t}$ for simplicity.

\begin{ap_definition}[Lipschitz Continuity]
\label{def1}
A function $f(x)$ is Lipschitz continuous on the set  $\Omega$, if there exists a constant $\epsilon>0$, $\forall x_{1}, x_{2}\in \Omega$ such that 
	\begin{equation}
		\label{eq:3.9}
		\|f(x_{1})-f(x_{2})\| \leq \epsilon\|x_{1}-x_{2}\|,
	\end{equation}
where $\epsilon$ is termed as the Lipschitz constant.
\end{ap_definition}  
Namely, the objective function $\mathcal{L}$ of TELL is Lipschitz continuous \textit{i.i.f.} $\|\nabla \mathcal{L}_{t}\|\leq \epsilon$. In other words, to meet the Lipschitz continuity, we need to prove that the upper boundary of $\nabla \mathcal{L}_{t}/\tau$ exists. To this end, we propose the following theorems.  

\begin{figure}[!t]
\centering
\includegraphics[width=0.58\textwidth]{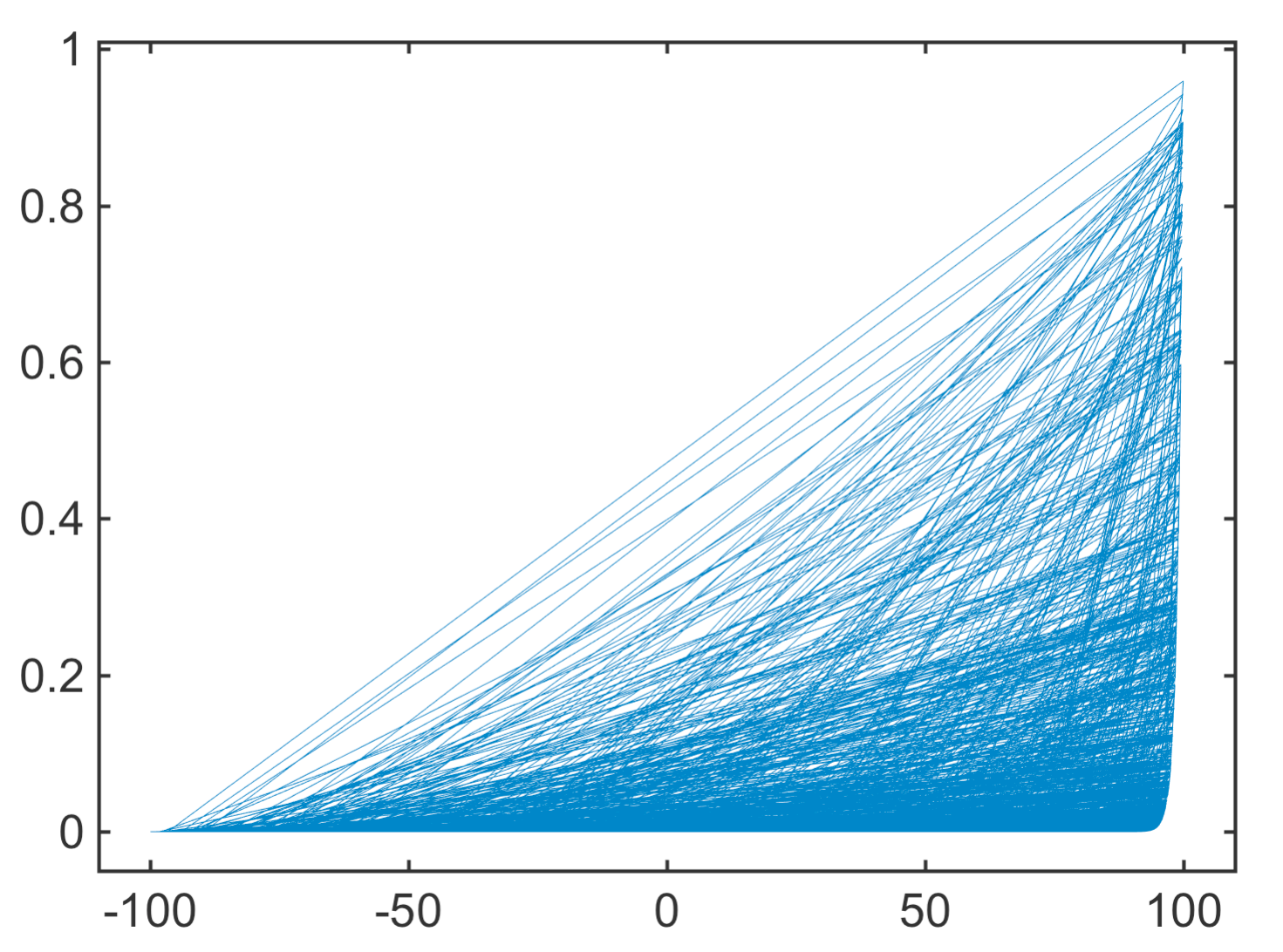}
\caption{\label{fig2} A toy example to show the bound of our loss function in the 1-dimensional case. The x-axis denotes the data points ($\bz$) randomly sampled from -100 to 100, and the y-axis denotes the corresponding loss value. One could see that our loss function will be bounded if $\bz$ is bounded.}
\end{figure}

\begin{theorem}
\label{lem1}
There exists $\epsilon>0$ such that $\|\nabla \mathcal{L}_{t}\|\leq \epsilon$, where $\epsilon = \tau+2\tau\max(\|\bz_{i}\|)$ 
	and 
	$\bz_{i}=\bW_{i}^{\top}\bx/\tau$.
\end{theorem}
\begin{proof}
Without loss of generality, we consider our loss in the form of 
\begin{equation}
\label{lem:eq1}
	\mathcal{L}(\bW_{i})=-\frac{\exp((\bW_{i}^{\top}\bx-2)/\tau)(\bW_{i}^{\top}\bx-2)}{\sum_{k}\exp((\bW_{k}^{\top}\bx-2)/\tau)}.
\end{equation}

Let $\bz_{i}=(\bW_{i}^{\top}\bx-2)/\tau$, we have

\begin{equation}
\label{lem:eq2}
	f(\bz_{i})=-\tau\frac{\exp(\bz_{i})(\bz_{i})}{\sum_{j}\exp(\bz_{j})}=-\tau\bp_{i}\bz_{i},
\end{equation}	
and then
\begin{align}
	\label{lem:eq3}
	\nabla_{i}f(\bz_{i})
		=&-\tau\left(\frac{(\exp(\bz_{i})+\exp(\bz_{i})\bz_{i})\sum_{j}\exp(\bz_{j})}
		{(\sum_{j}\exp(\bz_{j}))^{2}}-\frac{\exp(\bz_{i})\exp(\bz_{i})\bz_{i}}
		{(\sum_{j}\exp(\bz_{j}))^{2}}\right)\notag\\
		=&\tau(-\bp_{i}-\bp_{i}\bz_{i}+\bp_{i}\bp_{i}\bz_{i}).
\end{align}

As $0\leq \|\bp_{i}\|\leq 1$, we further have
\begin{align}
	\label{lem:eq4}
	\|\nabla_{i}f(\bz_{i}) \|&\leq \tau\|\bp_{i}\|(1+\|\bz_{i}\|+\bp_{i}\|\bz_{i}\|)
	\leq \tau(1+\|\bz_{i}\|+\|\bz_{i}\|).
\end{align}

It shows that our objective function $\cL(\bW_i)$ will be upper bounded by a positive real number $\epsilon$ when $\|\bz_{i}\|$ is bounded (see \figurename~\ref{fig2} for an illustrative example). In fact,  there exists the upper boundary of $\|\bz_{i}\|$ for any real-world data set. Moreover, without loss of generality, one could enforce $\|\bX_{i}\|=1$ and $\|\bOmega_{i}\|=1$ and thus $\|\bW_i\|=2$ is bounded.
\end{proof}

Based on Theorem~\ref{lem1}, we have the following convergence result based on~\citep{Boyd2003:subgradient}.
\begin{theorem}
	\label{thm2}
	One could always find an optimal model $\cL_{T}^{\ast}$ which is sufficiently close to the desired $\cL^{\ast}$ after $T$ steps, \textit{i.e.},
	\begin{equation}
		\label{thm2:eq1}
		\cL_{T}^{\ast}-\cL^{\ast}\leq \frac{\|\bW_{1}-\bW^{\ast}\|_{F}^{2}+\epsilon^{2}\sum_{t}^{T}\eta_{t}^{2}}{2\sum_{t=1}^{T}\eta_{t}}.
	\end{equation}
\end{theorem}
	
\begin{proof}
	Let $\bW^{\ast}=2\bOmega^{\ast}$ be the minimizer to our objective function, \textit{i.e.} Eq.~(\ref{eq3.3}), then
\begin{align}
	\label{thm1:eq3}
	\|\bW_{T+1}-\bW^{\ast}\|_{F}^{2}
	=&\|\bW_{T}-\bW^{\ast}\|_{F}^{2}
	-2tr(\eta_{T}\nabla\mathcal{L}_{T}^{\top}(\bW_{T}-\bW^{\ast}) )+\eta_{t}^{2}\|\nabla\mathcal{L}_{T} \|_{F}^{2},
\end{align}
where $tr(\cdot)$ denotes the trace of a matrix.

By recursively applying the above equation, we have
\begin{align}
	\label{thm2:eq4}
	\|\bW_{T+1}-\bW^{\ast}\|_{F}^{2}
	=&\|\bW_{1}-\bW^{\ast}\|_{F}^{2}-2\sum_{t=1}^{T}\eta_{t}tr(\bW_{t}-\bW^{\ast})+\sum_{t=1}^{T}\eta_{t}^{2}\|\nabla\cL_{t} \|_{F}^{2}.
\end{align}

As $\cL(\bW)$ satisfies the Lipschitz Continuity and according to the definition of gradient, \textit{i.e.}
\begin{equation}
	\label{thm2:eq5}
	f(x^{\ast})\geq f(x_{t})+\nabla\cL_{t}^{\top}(x^{\ast}-x_{t})
\end{equation}
then,
\begin{align}
	\label{thm2:eq6}
	\|\bW_{T+1}-\bW^{\ast} \|_{F}^{2}\leq & \|\bW_{1}-\bW^{\ast}\|_{F}^{2}-2\sum_{t=1}^{T}\eta_{t}(\cL_{t}-\cL^{\ast})+\epsilon^{2}\sum_{t=1}^{T}\eta_{t}^{2},
\end{align}

\begin{equation}
	\label{thm2:eq7}
	2\sum_{t=1}^{T}\eta_{t}(\cL_{t}-\cL^{\ast})\leq\|\bW_{1}-\bW^{\ast}\|_{F}^{2}+\epsilon^{2}\sum_{t=1}^{T}\eta_{t}^{2},
\end{equation}

\begin{equation}
	\label{thm2:eq8}
	\cL_{t}-\cL^{\ast}\geq \min_{t=1,2,\cdots,T}(\cL_{t}- \cL^{\ast})=\cL_{T}^{\ast} - \cL^{\ast},
\end{equation}
where $\cL_{T}^{\ast}$ is the best $\cL$ found so far in $T$ steps.

Combining Eq.~(\ref{thm2:eq7}) and Eq.~(\ref{thm2:eq8}), we finally have
\begin{equation}
	\label{thm2:eq9}
	\cL_{T}^{\ast}-\cL^{\ast}\leq \frac{\|\bW_{1}-\bW^{\ast}\|_{F}^{2}+\epsilon^{2}\sum_{t=1}^{T}\eta_{t}^{2}}{2\sum_{t=1}^{T}\eta_{t}}.
\end{equation}
\end{proof}

Based on Theorem~\ref{thm2}, the following two lemmas could be derived. 

\begin{ap_lemma}
	\label{lem1}
	For the fixed step size (\textit{i.e.} $\eta_{t}=\eta$) and $T\rightarrow\infty$, 
\begin{equation}
	\label{lem1:eq1}
	\mathcal{L}^{\ast}_{T}-\mathcal{L}^{\ast}\rightarrow\frac{\eta \epsilon^{2}}{2}.
\end{equation} 
\end{ap_lemma}
\begin{proof}
After $T$ steps, we have
\begin{align}
	\cL_{T}^{\ast}-\cL^{\ast}
	&\leq \frac{\|\bW_{1}-\bW^{\ast}\|_{F}^{2}+T\epsilon^{2}\eta^{2}}{2T\eta}
	=\frac{\|\bW_{1}-\bW^{\ast}\|_{F}^{2}/(T\eta)+\eta\epsilon^{2}}{2}.
\end{align}
\end{proof}

\begin{ap_lemma}
	\label{lem2}
	For the fixed step length (\textit{i.e.} $\eta_{t}=\eta/\nabla\mathcal{L}_{t}$) and $T\rightarrow\infty$, 
\begin{equation}
	\mathcal{L}^{\ast}_{T}-\mathcal{L}^{\ast}\rightarrow\frac{\eta \epsilon}{2}
\end{equation} 
\end{ap_lemma}
\begin{proof}
	Similar to the proof for Lemma~\ref{lem1}.
\end{proof}

Lemma~\ref{lem1}---\ref{lem2} show that the loss will eventually converge to $\mathcal{L}^{\ast}$ with a radius of $\frac{\eta\epsilon^{2}}{2}$ and $\frac{\eta \epsilon}{2}$ within $T$ steps.

\newpage
\renewcommand{\thetable}{S\arabic{table}}

\section*{Appendix B.}
As a neural surrogate of the vanilla $k$-means, the performance of TELL naturally relates to the network initialization. Just like all machine learning methods, TELL faces the problem of model selection as well. To solve model selection in unsupervised setting, motivated by the vanilla $k$-means, in practice, we run TELL several times with different random initializations and obtain the final result by the run with the minimal clustering loss $\cL_{clu}$. To show the correctness of such a criterion, we provide the clustering performance of TELL in five different runs on MNIST and CIFAR-10 in Table~\ref{tab:mnist_app} and \ref{tab:cifar_app}.

\setcounter{table}{0}
\begin{table}[!t]
\centering
\begin{tabular}{c|c|c|c|c}
\hline
Run & ACC    & NMI    & ARI    & $\cL_{clu}$ \\ \hline
1    & \textbf{95.16} & \underline{88.83} & \textbf{89.66} & \textbf{0.018138861}   \\ \hline
2    & \underline{95.10} & \textbf{89.03} & \underline{89.53} & \underline{0.018156468}   \\ \hline
3    & 82.88 & 87.00 & 81.01 & 0.018166682   \\ \hline
4    & 82.04 & 83.90 & 77.05 & 0.018166506   \\ \hline
5    & 81.69 & 83.87 & 77.33 & 0.018164165   \\ \hline
\end{tabular}%
\caption{The clustering performance on MNIST in five different runs with random initialization. The best and second-best results are shown in bold and underline, respectively.}
\label{tab:mnist_app}
\end{table}

\begin{table}[!t]
\centering
\begin{tabular}{c|c|c|c|c}
\hline
Run & ACC    & NMI    & ARI    & $\cL_{clu}$ \\ \hline
1    & 24.44 & \textbf{10.59} & 5.37 & 0.018088905   \\ \hline
2    & \underline{24.97} & 10.19 & \underline{5.73} & \underline{0.018088135}   \\ \hline
3    & 22.19 & 10.40 & 5.50 & 0.018093264   \\ \hline
4    & 23.25 & 10.12 & 5.36 & 0.018089087   \\ \hline
5    & \textbf{25.65} & \underline{10.41} & \textbf{5.96} & \textbf{0.018088113}   \\ \hline
\end{tabular}%
\caption{The clustering performance on CIFAR-10 in five different runs with random initialization. The best and second-best results are shown in bold and underline, respectively.}
\label{tab:cifar_app}
\end{table}

The results show that the clustering accuracy of TELL is correlated to the clustering loss $\cL_{clu}$. On both data sets, the best and second-best results correspond to the smallest two clustering losses. As the clustering loss measures the within-cluster distance, a smaller clustering loss indicates a more compact clustering result. Thus, in practice, one could always run TELL several times and choose the run with the smallest $\cL_{clu}$ as the final output.

\vskip 0.2in
\bibliography{reference}

\end{document}